\documentclass[10pt,twocolumn,letterpaper]{article}

\usepackage{iccv}
\usepackage{times}
\usepackage{epsfig}
\usepackage{graphicx}
\usepackage{amsmath}
\usepackage{amssymb}

\usepackage{multirow}

\newcommand{\bmx}[0]{\begin{bmatrix}}
\newcommand{\emx}[0]{\end{bmatrix}}

\newcommand{\vect}[1]{\mathbf{#1}}
\newcommand{\vects}[1]{\boldsymbol{#1}}
\newcommand{\matr}[1]{\mathbf{#1}}

\newcommand{\vb}[0]{\vect{b}}
\newcommand{\vd}[0]{\vect{d}}
\newcommand{\vc}[0]{\vect{c}}
\newcommand{\vo}[0]{\vect{o}}

\newcommand{\vh}[0]{\vect{h}}
\newcommand{\vi}[0]{\vect{i}}
\newcommand{\vv}[0]{\vect{v}}
\newcommand{\vx}[0]{\vect{x}}
\newcommand{\vw}[0]{\vect{w}}
\newcommand{\vp}[0]{\vect{p}}
\newcommand{\vf}[0]{\vect{f}}

\newcommand{\mE}[0]{\matr{E}}
\newcommand{\mW}[0]{\matr{W}}

\newcommand{\mU}[0]{\matr{U}}

\newcommand{\mA}{\matr{A}}

\newcommand{\TT}[0]{\vects{\theta}}

\usepackage[breaklinks=true,bookmarks=false]{hyperref}

\iccvfinalcopy % *** Uncomment this line for the final submission

\def\iccvPaperID{913} % *** Enter the ICCV Paper ID here

\ificcvfinal\fi
\begin{document}

%%%%%%%%% TITLE
{
\title{Describing Videos by Exploiting Temporal Structure}

\author{
Li Yao \\
{Universit\'{e} de Montr\'{e}al}\\
{\tt\small li.yao@umontreal.ca} \\
\and
Atousa Torabi \\
{Universit\'{e} de Montr\'{e}al}\\
{\tt\small atousa.torabi@umontreal.ca} \\
\and
Kyunghyun Cho \\ 
{Universit\'{e} de Montr\'{e}al}\\
{\tt\small kyunghyun.cho@umontreal.ca} \\
\and
Nicolas Ballas \\
{Universit\'{e} de Montr\'{e}al}\\
{\tt\small nicolas.ballas@umontreal.ca} \\
\and
Christopher Pal \\
{ \'Ecole Polytechnique de Montr\'{e}al}\\
{\tt\small christopher.pal@polymtl.ca} \\
\and
Hugo Larochelle \\
{Universit\'{e} de Sherbrooke} \\
{\tt\small hugo.larochelle@usherbrooke.ca}\\ 
\and
Aaron Courville \\
{Universit\'{e} de Montr\'{e}al}\\
{\tt\small aaron.courville@umontreal.ca} \\
}
}

\maketitle
%\thispagestyle{empty}

%%%%%%%%% ABSTRACT
\begin{abstract}

%% Blah blah
Recent progress in using recurrent neural networks (RNNs) for image description has motivated the
exploration of their application for video description. However, while images are static, working 
with videos requires modeling their dynamic temporal structure and then properly integrating
that information into a natural language description. 
In this context, we propose an approach that successfully takes into account 
both the local and global temporal structure of videos to produce descriptions.
First, our approach incorporates a spatial temporal 3-D convolutional neural network (3-D CNN) representation
of the short temporal dynamics. The 3-D CNN representation is trained on video action recognition tasks,
so as to produce a representation that is tuned to human motion and behavior.
Second we propose a temporal attention mechanism that
allows to go beyond local temporal modeling and \emph{learns} to automatically select the most relevant
temporal segments given the text-generating RNN. 
Our approach exceeds the current state-of-art 
for both BLEU and METEOR metrics on the Youtube2Text dataset.
We also present results on a new, larger and more challenging dataset of paired 
video and natural language
descriptions. 
\end{abstract}

%%%%%%%%% BODY TEXT
 
%%%%%%%%% BODY TEXT
\section{Introduction}

%\alert{Litterature review does not have separate section?}

%The ability to automatically generate natural language descriptions of
%video content has the potential to dramatically improve content-based information retrieval.

%% Recurrent neural networks (RNNs) have recently enjoyed a resurgence in
%% popularity due to a number of high profile successes in natural language
%% processing applications. In particular, there has recently been strong interest in
%% using RNN-based models for the automatic generation of image
%% descriptions. Most of these models use input image features derived from
%% large convolutional neural networks (CNNs) trained on the large Imagenet
%% dataset as an input to a long short-term memory (LSTM) network natural
%% language decoder that generates natural text output conditional on the
%% image features.

%% In this paper we apply the combination of RNNs and CNNs
%% to a related task: the automatic generation of \emph{video} descriptions.
The task of automatically describing videos containing rich and open-domain activities 
poses an important challenges for computer vision and
machine learning research. 
It also has a variety of practical
applications. For example, every minute, 100 hours of video are uploaded to
YouTube.\footnote{\url{https://www.youtube.com/yt/press/statistics.html} accessed on 2015-02-06.}
However, if a video is poorly tagged, its utility is dramatically diminished~\cite{morsillo2010youtube}.
Automatic video description generation has the potential to help improve
indexing and search quality for online videos.  In conjunction with speech synthesis technology, annotating video with
natural language descriptions also has the potential to benefit the visually
impaired.

%% In recent years descriptive video, also known as audio description, has become more widely available on many digital medial products such as DVDs, due in large part to an organization and technology known as Descriptive Video Services (DVS). The technique consists of including an additional audio track with a concise description of the visual content in a scene, spoken by a narrator. % during select segments of the video.
%%  The goal of DVS is to provide a richer experience for the visually impaired. Automatic video description could further democratize this technology.

\begin{figure}[t]
\center
\includegraphics[width=8cm]{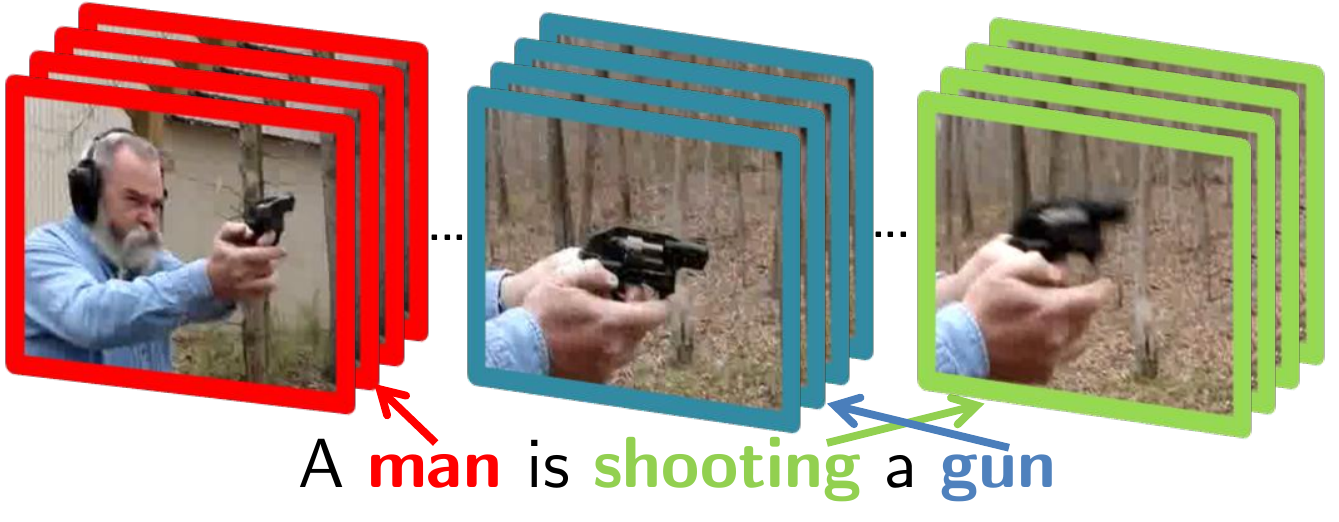}
\caption{High-level visualization of our approach to video description generation. We incorporate
models of both the local temporal dynamic (i.e.\ within blocks of a few frames) of videos, as well as their global temporal structure. The local structure is modeled using the temporal feature maps of a 3-D CNN, while a temporal attention mechanism is used to combine information across the entire video. For each generated
word, the model can focus on different temporal regions in the video. 
For simplicity, we highlight only the region having the maximum attention above.}
\label{fig:motiv}
\end{figure}

While image description generation is already considered a very challenging
task, the automatic generation of video description carries additional
difficulties. Simply dealing with the sheer quantity of information contained in video data is one
such challenge. Moreover, video description involves generating a sentence
to characterize a video clip lasting typically 5 to 10 seconds, or 120 to 240
frames.  Often such clips contain complex interactions of actors and objects
that evolve over time. All together it amounts to a vast quantity of information, and
attempting to represent this information using a single, temporally collapsed feature
representation is likely to be prone to clutter, with temporally distinct events and objects being
potentially fused incoherently. It is therefore important that an automatic
video description generator {\it exploit the temporal structure} underlying video.

We argue that there are two categories of temporal structure present in
video: (1) local structure and (2) global structure. Local temporal
structure refers to the fine-grained motion information that characterizes
punctuated actions such as ``answering the telephone'' or ``standing
up''. Actions such as these are relatively localized in time, evolving over
only a few consecutive frames. 
On the other hand, when we refer to global temporal structure in video, 
we refer to the sequence in which objects, actions, scenes and people, etc. appear in a video.
%On the other hand, global temporal structure in video refers to the
%sequence of such actions that combine to make up the video. 
Video description may well be termed video summarization, because we
typically look for a single sentence to summarize what can be a rather
elaborate sequence of events. Just as good image descriptions often focus
on the more salient parts of the image for description, we argue that good video
description systems should selectively focus on the most salient features
of a video sequence.
% -- in order to avoid fusing temporally disparate video features into an incoherent description.

Recently, \mbox{Venugopalan et al.~\cite{venugopalan2014translating}} used a
so-called encoder--decoder neural network framework~\cite{Cho2014} to
automatically generate the description of a video clip. They extracted
appearance features from each frame of an input video clip using a previously trained
convolutional neural network~\cite{krizhevsky2012}. The features from all
the frames, or subsampled frames, were then collapsed via simple averaging to result in
a single vector representation of the entire video clip. Due to this indiscriminate
averaging of all the frames, this approach risks ignoring much of the temporal
structure underlying the video clip. For instance, it is not possible to tell
the order of the appearances of two objects from the collapsed features.

In this paper, we introduce a temporal attention mechanism to
exploit \emph{global} temporal structure. We also augment the appearance features
with action features that encode \emph{local} temporal structure. Our
action features are derived from a spatio-temporal convolutional neural network
(3-D CNN)~\cite{tran2014c3d, karpathy2014large, ji2013}. The temporal attention mechanism is based on
a recently proposed soft-alignment method~\cite{Bahdanau2014} which was used
successfully in the context of machine translation. While generating a
description, the temporal attention mechanism selectively focuses on a small
subset of frames, making it possible for the generator to describe only the
objects and/or activities in that subset (see Fig.~\ref{fig:motiv} for the
graphical illustration). Our 3-D CNN, 
on the other hand, starts from both temporally and spatially local motion descriptors of
video and hierarchically extracts more abstract action-related
features. These features preserve and emphasize important local structure
embedded in video for use by the description generator.

We evaluate the effectiveness of the proposed mechanisms for exploiting temporal
structure on the most widely used open-domain video description dataset, called
the Youtube2Text dataset~\cite{Chen2011}, which consists of 1,970 video clips with
multiple descriptions per video. We also test the proposed approaches on
a much larger, and more recently proposed, dataset based on the descriptive
video service (DVS) tracks in DVD movies~\cite{Torabi2015}, which contains
49,000 video clips.

% CP - We need an explicit 'this is our contributions' paragraph, here is a draft merged into the previous last paragraph of the intro
Our work makes the following contributions: 1) We propose the use of a
novel 3-D CNN-RNN encoder-decoder architecture which captures local spatio-temporal information. We find that
despite the promising results generated by both prior work and our own here using static frame 
CNN-RNN video description methods, our experiments suggest that it is
indeed important to exploit local temporal
structure when generating a description of video. 
%
%Our experiments suggest that it is indeed important to exploit temporal
%structure when generating a description of video. 
% CP- The statement above might lead reviewers to say 'of course', so I've added 
% the qualification beforehand which I think was motivating this statement
2) We propose the use of an attention mechanism within a CNN-RNN encoder-decoder framework for video description and we demonstrate through our experiments that it allows features obtained through the global analysis of static frames throughout the video to be used more effectively for video description generation. Furthermore, 3) we observe that
the improvements brought by exploiting global and local temporal information are
complimentary, with the best performance achieved when both the
temporal attention mechanism and the 3-D CNN are used together.

\section{Video Description Generation Using an Encoder--Decoder Framework}

In this section, we describe a general approach, based purely on neural networks
to generate video descriptions. This approach is based on the
encoder-decoder framework~\cite{Cho2014}, which has been successfully used in
machine translation~\cite{SutskeverI2014,Cho2014,Bahdanau2014} as well as image
caption generation~\cite{Kiros2014,donahue2014long,vinyals2014show,xu2015show,KarpathyCVPR14}.

\subsection{Encoder-Decoder Framework}

The encoder-decoder framework consists of two neural networks; the encoder and
the decoder. The encoder network $\phi$ encodes the input $\vx$ into a
continuous-space representation which may be 
%either a single vector $\vv$ with
%fixed-dimensionality or 
a variable-sized set $V=\left\{ \vv_1, \dots, \vv_n \right\}$
of continuous vectors:
\begin{align*}
    V=\left\{ \vv_1, \dots, \vv_n \right\} = \phi(\vx).
\end{align*}
%In the case of a single vector output, we set $n=1$.

The architecture choice for the encoder $\phi$ depends on the type of input.
For example, in the case of machine translation, it is natural to use a
recurrent neural network (RNN) for the encoder, since the input is a variable-length
sequence of symbols~\cite{SutskeverI2014,Cho2014}. With an
image as input, a convolutional neural network (CNN) is another good alternative~\cite{xu2015show}.

The decoder network generates the corresponding output $y$ from the encoder
representation $V$. As was the case with the encoder, the decoder's architecture
must be chosen according to the type of the output. When the output is a natural
language sentence, which is the case in automatic video description, an RNN is
a method of choice. 

The decoder RNN $\psi$ runs sequentially over the output sequence.  In brief, to
generate an output $y$, at each step $t$ the RNN updates its internal state
$\vh_t$ based on its previous internal state $\vh_{t-1}$ as well as the previous
output $y_{t-1}$ and the encoder representation $V$, and then outputs a symbol
$y_t$:
\begin{align}
    \label{eq:rnn}
    \left[
        \begin{array}{c}
            y_t \\
            \vh_t
        \end{array}
    \right]
    = \psi(\vh_{t-1}, y_{t-1}, V)
\end{align}
where for now we simply note as $\psi$ the function updating the RNN's internal
state and computing its output.  The RNN is run recursively until the
end-of-sequence symbol is generated, i.e., $y_t=\left<\text{eos}\right>$.

In the remaining of this section, we detail choices for the encoder
and decoder for a basic automatic video description system, taken from 
\cite{venugopalan2014translating} and on which our work builds.

\subsection{Encoder: Convolutional Neural Network}
\label{sec:cnn}

Deep convolutional neural networks (CNNs) have recently been successful at
large-scale object recognition~\cite{krizhevsky2012,szegedyL}.
Beyond the object recognition task itself, CNNs trained for object recognition
have been found to be useful in a variety of other computer vision tasks such as
object localization and detection (see, e.g., \cite{Sermanet14}). This has opened a door to
a flood of computer vision systems that exploit representations from upper
or intermediate layers of a CNN as generic high-level features for vision. 
%One such example is image caption generation where the pretrained CNN is
%used as an encoder together with a decoder based on a recurrent neural network
%(RNN)~\cite{Kiros2014,donahue2014long,vinyals2014show,xu2015show,KarpathyCVPR14}
%Indeed, once a CNN is trained to recognize an object in an input image, we can use the
%activation of the intermediate layers as the representation of the image. 
For instance, the activation of the last fully-connected layer can be used as a
fixed-size vector representation~\cite{Kiros2014}, or the feature map of the
last convolutional layer can be used as a set of spatial feature
vectors~\cite{xu2015show}.

%\alert{TBD: Briefly describe a convolutional neural network. Is this necessary?
%CP: More important to just say precisely which CNN we use here more clearly.}
%Cho: I agree with CP, and we specify which CNN we use in the experiment.

In the case where the input is a video clip, an image-trained CNN can be used for
each frame separately, resulting in a single vector representation $\vv_i$ of the
$i$-th frame. 
%Collectively, the CNN-based encoder returns a set $V=\left\{ v_i
%\right\}_{i=1}^n$ of $n$ vectors corresponding to the $n$ frames in the video.
% CPAL added this
This is the approach proposed by \cite{venugopalan2014translating}, which used
the convolutional neural network from \cite{krizhevsky2012}. In our work here,
we will also consider using the CNN from \cite{szegedyL}, which has demonstrated
higher performance for object recognition. 
%
% CPAL - removed this compressed version as it is covered in the next section
% Unlike our approach, Venugopalan et al. also collapse the vector representations from a CNN for frames in a video clip into a single vector through averaging.

%In \cite{venugopalan2014translating}, Venugopalan et al. used the CNN to extract
%a vector representation of each frame of a video clip. They collapse all these
%frame-wise vector representations into a single vector by simply averaging them,
%and this single vector is considered as an encoded representation $h$ of the
%input video clip. The decoder subsequently generates a description from this
%representation. We emphasize here that this strategy of simple averaging
%effectively ignores any temporal relationships among the frames, leading to the
%loss of any temporal structure in the video clip.

\subsection{Decoder: Long Short-Term Memory Network}
\label{sec:decoder}

As discussed earlier, it is natural to use a recurrent neural network (RNN) as a
decoder when the output is a natural language sentence. This has been
empirically confirmed in the contexts of machine
translation~\cite{SutskeverI2014,Cho2014,Bahdanau2014}, image caption
generation~\cite{vinyals2014show,xu2015show} and video description
generation in open~\cite{venugopalan2014translating} and 
closed~\cite{donahue2014long} domains. Among these recently successful
applications of the RNN in natural language generation, it is noticeable that
most of
them~\cite{SutskeverI2014,Cho2014,Bahdanau2014,vinyals2014show,xu2015show}, if
not all, used long short-term memory (LSTM)
units~\cite{Hochreiter+Schmidhuber-1997} or their variant, gated recurrent units
(GRU)~\cite{Cho2014}.  In this
paper, we also use a variant of the LSTM units, introduced in \cite{Zaremba2014},
as the decoder.

The LSTM decoder maintains an internal memory state $\vc_t$ in addition to the
usual hidden state $\vh_t$ of an RNN (see Eq.~\eqref{eq:rnn}). The hidden state
$\vh_t$ is the memory state $\vc_t$ modulated by an output gate:
\begin{align*}
    \vh_t = \vo_t \odot \vc_t,
\end{align*}
where $\odot$ is an element-wise multiplication. The output gate $\vo_t$ is computed by
\begin{align*}
    \vo_t = \sigma(\mW_o \mE\left[y_{t-1}\right] + \mU_o \vh_{t-1} + \mA_o \varphi_t(V) + \vb_o),
\end{align*}
where $\sigma$ is the element-wise logistic sigmoid function and $\varphi_t$ is a
time-dependent transformation function on the encoder features. $\mW_o$, $\mU_o$, $\mA_o$ and
$\vb_o$ are, in order, the weight matrices for the input, the previous hidden
state, the context from the encoder and the bias. $\mE$ is a word embedding
matrix, and we denote by $\mE\left[y_{t-1}\right]$ an embedding vector of word
$y_{t-1}$.

The memory state $\vc_t$ is computed as a weighted sum between the previous memory
state $\vc_{t-1}$ and the new memory content update $\tilde{\vc}_t$:
\begin{align*}
    \vc_t = \vf_t \odot \vc_{t-1} + \vi_t \odot \tilde{\vc}_t,
\end{align*}
where the coefficients -- called forget and input gates respectively --
are given by
\begin{align*}
    \vf_t = \sigma(\mW_f \mE\left[y_{t-1}\right] + \mU_f \vh_{t-1} + \mA_f \varphi_t(V) + \vb_f), \\
    \vi_t = \sigma(\mW_i \mE\left[y_{t-1}\right] + \mU_i \vh_{t-1} + \mA_i \varphi_t(V) + \vb_i).
\end{align*}
The updated memory content $\tilde{\vc}_t$ also depends on the current input
$y_{t-1}$, previous hidden state $\vh_{t-1}$ and the features from the encoder
representation $\varphi_t(V)$:
\begin{align*}
    \tilde{\vc}_t = \tanh(\mW_c \mE\left[y_{t-1}\right] + \mU_c \vh_{t-1} + \mA_c \varphi_t(V) + \vb_c).
\end{align*}

Once the new hidden state $\vh_t$ is computed, a probability distribution
over the set of possible words is obtained using a 
single hidden layer neural network
\begin{align}
    \vp_t = {\rm softmax}(&\mU_p \tanh(\mW_p [\vh_t,\varphi_t(V),\mE\left[y_{t-1}\right]] + \vb_p) + \vd),
\end{align}
where $\mW_p, \mU_p, \vb_p, \vd$ are the parameters of this
network, $[\dots]$ denotes vector concatenation. The softmax
function allows us to interpret $\vp_t$ as the probabilities
of the distribution $p(y_t \mid y_{<t}, V)$ over words.

At a higher level, the LSTM decoder can be written down as 
\begin{align}
    \label{eq:lstm}
    \left[
        \begin{array}{c}
            p(y_t\mid y_{< t}, V) \\
            \vh_t \\
            \vc_t
        \end{array}
    \right]
    = \psi(\vh_{t-1}, \vc_{t-1}, y_{t-1}, V).
\end{align}

It is then trivial to generate a sentence from the LSTM decoder. For instance,
one can recursively evaluate $\psi$ and sample from the returned $p(y_t \mid
\dots)$ until the sampled $y_t$ is the end-of-sequence symbol. One can also
approximately find the sentence with the highest probability by
using a simple beam search~\cite{SutskeverI2014}.

In \cite{venugopalan2014translating}, Venugopalan et al. used this type of LSTM
decoder for automatic video description generation. However, in their work the feature
transformation function $\varphi_t$ consisted in a simple averaging, i.e., 
\begin{align}
    \label{eq:simple_avg}
    \varphi_t(V) = \frac{1}{n} \sum_{i=1}^n \vv_i,
\end{align}
where the $v_i$'s are the elements of the set $V$ returned by the CNN encoder from
Sec.~\ref{sec:cnn}. This averaging
effectively collapses all the frames, indiscriminate of their temporal
relationships, leading to the loss of temporal structure underlying the input
video.

\section{Exploiting Temporal Structure in Video Description Generation}
\label{sec:contrib}

In this section, we delve into the main contributions of this paper and propose 
an approach for exploiting both the local and global temporal structure
in automatic video description. 
%model proposed earlier in \cite{venugopalan2014translating}.
%AC: cut the above to make our contribution sound less marginal.

\subsection{Exploiting Local Structure: \\ A Spatio-Temporal Convolutional Neural Net}
\label{sec:3dconv}

We propose to model the local temporal structure of videos at the level
of the temporal features $V=\left\{ \vv_1, \dots, \vv_n \right\}$ that are extracted
by the encoder. 
Specifically, we propose to use a spatio-temporal convolutional neural network
(3-D CNN) which has recently been demonstrated to capture well the temporal
dynamics in video clips~\cite{tran2014c3d,karpathy2014large}.

We use a 3-D CNN to build the higher-level representations that preserve
and summarize the local motion descriptors of short frame sequences. This is done
by first dividing the input video clip into a 3-D spatio-temporal grid of
$16\times 12 \times 2$ (width $\times$ height $\times$ timesteps) cuboids.
Each cuboid is represented by concatenating the histograms of oriented
gradients, oriented flow and motion boundary (HoG, HoF,
MbH)~\cite{dalal2006,WangH2009} with 33 bins. This transformation is done in
order to make sure that local temporal structure (motion features) are well
extracted and to reduce the computation of the subsequence 3-D CNN.

Our 3-D CNN architecture is composed of three 3-D convolutional layer, each followed by rectified linear
activations (ReLU) and local max-pooling. From the activation of the
last 3-D convolution+ReLU+pooling layer, which preserves the temporal
arrangement of the input video and abstracts the local motion features, we can obtain 
a set of temporal feature
vectors by max-pooling along the spatial dimensions
(width and height) to get feature vectors that each summarize the content
over short frame sequences within the video. 
Finally, these feature vectors are combined, by concatenation, with the image
features extracted from single frames taken at similar positions across the video.
Fig.~\ref{fig:architecture} illustrates the complete architecture of the described 3-D CNN.
 Similarly to the object recognition trained CNN (see Sec.~\ref{sec:cnn}), the 3-D  CNN is pre-train on  activity recognition datasets.
%when used during training for activity recognition.

%% Move into supp-mat
%% Similarly to the object recognition trained CNN (see Sec.~\ref{sec:cnn}), we train our 3-D
%% CNN to perform activity recognition. We flatten the
%% activation of the last 3-D convolution+ReLu+pooling layer and feed it to a
%% linear classifier to compute the output probability.  We aggregate three
%% datasets, UCF101 (13,320 videos and 101 actions, \cite{Khurram2012}), HMDB51
%% (3,700 videos and 51 actions, \cite{Kuehne2011}) and a subset of Sports-1M
%% (50,000 videos from Youtube and 487 actions, \cite{karpathy2014large}), to do
%% multitask learning. 

%%  Only the first three convolutional layers
%% are ultimately used when extracting the local temporal features. 
%See the Supplementary Material for more details on how training was performed.

\begin{figure*}
\center
\begin{minipage}{0.65\textwidth}
\includegraphics[width=\columnwidth]{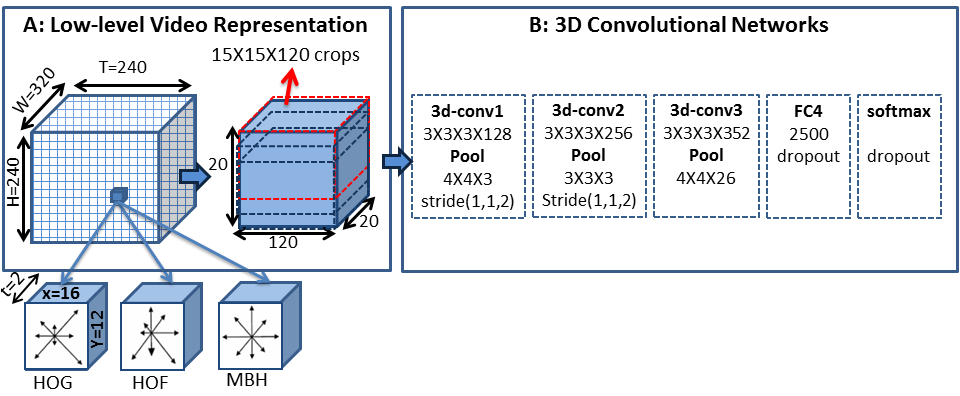}
\end{minipage}
\hspace{1mm}
\begin{minipage}{0.3\textwidth}
\caption{\small Illustration of the spatio-temporal convolutional neural
    network (3-D CNN).
    %In a 3DConv layer,
%convolution dimensions are $width \times height \times time \times channels$.
%For 3-D pooling dimensions are $width \times height \times time$. The stride
%dimensions for pooling of the first and second convolution layers are $width
%\times height \times time$. 
This network is trained for activity recognition. 
%(see Supplementary Material for details). 
Then, only the convolutional
layers are involved when generating video descriptions.
%In training, random crops (in red) are extracted and
%input to the first convolutional layer.
}
\label{fig:architecture}
\end{minipage}
\end{figure*}

\subsection{Exploiting Global Structure: \\ A Temporal Attention Mechanism}
\label{sec:attention}

\begin{figure}%[t]
\center
\includegraphics[width=7cm]{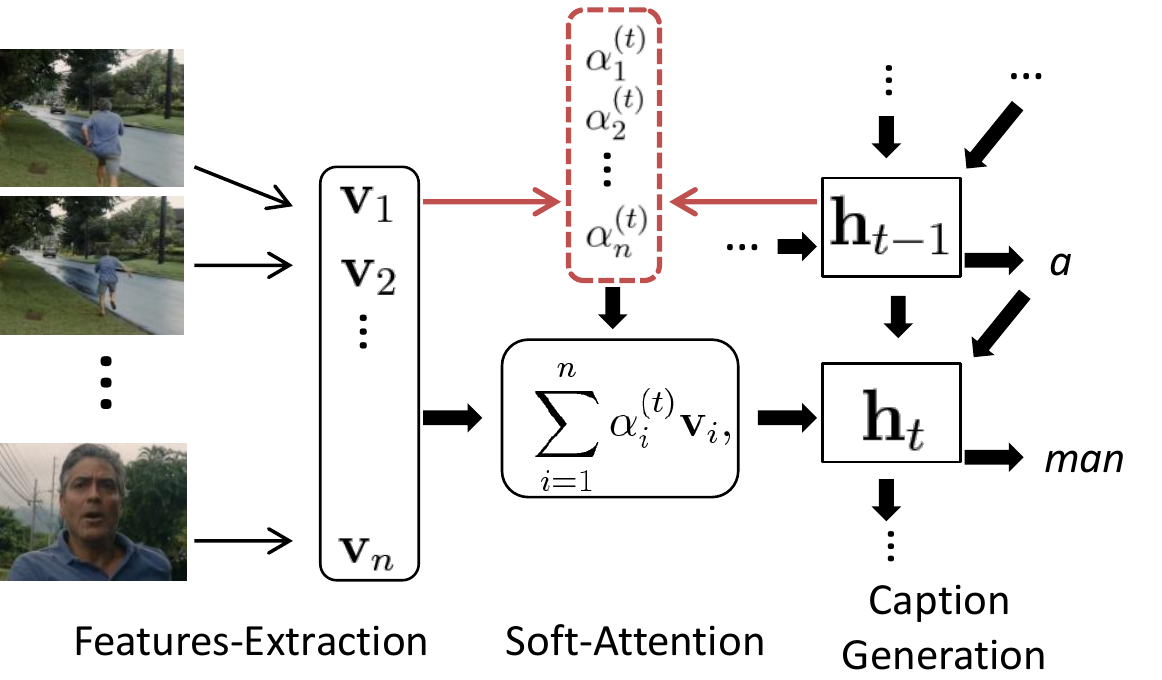}
\caption{Illustration of the proposed temporal attention mechanism in the LSTM decoder}
\label{fig:main}
\end{figure}

The 3-D CNN features of the previous section allows us to better represent
short-duration actions in a subset of consecutive frames. However, representing a complete video by averaging these local temporal features
as in Eq.~\ref{eq:simple_avg} would jeopardize 
the model's ability to exploit the video's global temporal structure.

Our approach to exploiting such non-local temporal structure is to
let the decoder selectively focus on only a small subset of frames at a time. By
considering subsets of frames in sequence, the model can exploit the temporal
ordering of objects and actions across the entire video clip and avoid
conflating temporally disparate events. Our approach also has the potential
of allowing the model to focus on key elements of the video that may have
short duration. Methods that collapse the temporal structure risk
overwhelming these short duration elements.

Specifically, we propose to adapt the recently proposed soft attention mechanism from
\cite{Bahdanau2014}, which allows the decoder to weight each temporal feature vector $V=\left\{ \vv_1, \dots, \vv_n \right\}$.
%\footnote{Note that the encoder in our definition is able to return a
%    {\it set of feature vectors} corresponding to, for instance, spatial or
%temporal dimensions of the input image or video.} 
This approach has been used
successfully by Xu et al.~\cite{xu2015show} for exploiting spatial structure
underlying an image. Here, we thus adapt it to exploit the temporal structure 
of video instead.

%We start with the set $V$ of frame-wise feature vectors from Sec.~\ref{sec:cnn}.
Instead of a simple averaging strategy (as shown in Eq.~\eqref{eq:simple_avg}), we
take the \emph{dynamic} weighted sum of the temporal feature vectors such that
\begin{align*}
    \varphi_t(V) = \sum_{i=1}^n \alpha_i^{(t)} \vv_i,
\end{align*}
where $\sum_{i=1}^n \alpha_i^{(t)} = 1$ and $\alpha_i^{(t)}$'s are computed at
each time step $t$ inside the LSTM decoder (see Sec.~\ref{sec:decoder}). We
refer to $\alpha_i^{(t)}$ as the attention weights at time $t$.

The attention weight $\alpha_i^{(t)}$ reflects the relevance of the $i$-th temporal feature in
the input video given all the previously generated words, i.e., $y_1, \dots
y_{t-1}$.  Hence, we design a function that takes as input the previous hidden
state $\vh_{t-1}$ of the LSTM decoder, which summarizes all the previously
generated words, and the feature vector of the $i$-th temporal feature and returns the
unnormalized relevance score $e_i^{(t)}$:
\begin{align*}
    e_i^{(t)} = \vw^\top \tanh \left(\mW_a \vh_{t-1} + \mU_a \vv_i + \vb_a \right),
\end{align*}
where $\vw$, $\mW_a$, $\mU_a$ and $\vb_a$ are the parameters that are estimated 
together with all the other parameters of the encoder and decoder networks.

Once the relevance scores $e_i^{(t)}$ for all the frames $i=1,\dots,n$ are
computed, we normalize them to obtain the $\alpha_i^{(t)}$'s:
\begin{align*}
    \alpha_i^{(t)} = \exp\left\{e_i^{(t)}\right\} / \sum_{j=1}^n \exp\left\{
        e_j^{(t)}\right\}.
\end{align*}
We refer to the {\it attention mechanism} as this whole process of computing the
unnormalized relevance scores and normalizing them to obtain the attention
weights.

The attention mechanism allows the decoder to selectively focus on only a
subset of frames by increasing the attention weights of the corresponding
temporal feature. However, we do not explicitly force this type of selective attention to
happen. Rather, this inclusion of the attention mechanism enables the decoder
to exploit the temporal structure, {\it if} there is useful temporal structure
in the data. Later in Sec.~\ref{sec:exp}, we empirically show that this is
indeed the case.
See Fig.~\ref{fig:main} for the graphical illustration of the temporal
attention mechanism.

\section{Related Work}

% \subsection{Video Description Generation}

Video description generation has been investigated and studied
in other work, such as ~\cite{kojima2002,barbu2012,rohrbach2013}. Most of these examples have,
however, constrained the domain of videos as well as the activities and objects
embedded in the video clips. Furthermore, they tend to rely on hand-crafted
visual representations of the video, to which template-based or shallow statistical
machine translation approaches were applied. In contrast, the approach we
take and propose in this paper aims at open-domain video description generation
with deep trainable models starting from low-level video representations,
including raw pixel intensities (see Sec.~\ref{sec:cnn}) and local motion
features (see Sec.~\ref{sec:3dconv}).

In this sense, the approach we use here is more closely related to the recently
introduced static image caption generation approaches based mainly on neural
networks~\cite{Kiros2014,donahue2014long,vinyals2014show,xu2015show,KarpathyCVPR14}.
A neural approach to static image caption generation has recently been applied
to video description generation by Venugopalan et
al.~\cite{venugopalan2014translating}. However, their direct adaptation of the
underlying static image caption generation mechanism to the videos is limited by
the fact that the model tends to ignore the temporal structure of the underlying
video.
Such structure has demonstrated to be helpful in the context of event and action classification~\cite{tang2012learning,gaidon2013temporal,bojanowski2014weakly}, and is explored in this paper.
Other recent work \cite{Rohrbach2015} has explored the use of DVS annotated
video for video description research and has underscored the observation that
DVS descriptions are typically much more relevant and accurate descriptions of
the visual content of a video compared to movie scripts. They present results
using both DVS and script based annotations as well as cooking activities. 
% Cho: maybe not ncessarily detailed.
%In
%their work they use a combination of features consisting of the improved dense
%trajectories of \cite{wang2013}, the large scale object detection convolutional
%neural network (CNN) from \cite{Hoffman2014}, and the scene classification CNNs
%from \cite{Zhou2014}. They then use Wordnet, semantic role labelling and word
%sense disambiguation to create a semantic representation in the form of
%(SUBJECT, VERB, OBJECT, LOCATION) tuples. A conditional random field (CRF) is
%used to predict the verb, object and location attributes and the resulting
%semantic representation is translated using an approach discussed in
%\cite{rohrbach2013}.
%
% removed lee2009convolutional to make it fit
%

While other work has explored 3-D Deep Networks for video \cite{taylor2010convolutional,ji2013,KarpathyCVPR14,simonyan2014} our particular approach differs in a number of ways from prior work in that it is based on CNNs as opposed to other 3-D deep architectures and we focus on pre-training the model on a number of widely used action recognition datasets. In contrast to other 3-D CNN formulations, the input to our 3-D CNN consists of features derived from a number of state of the art image descriptors. Our model is also fully 3-D in that we model entire volumes across a video clip.
%
%\subsection{Neural Video Modeling}
In this paper, we use a state-of-the-art static convolutional neural network (CNN) and
a novel spatio-temporal  3-D CNN to model input video clips. This way of
modeling video using feedforward convolutional neural networks, has become
increasingly popular
recently~\cite{venugopalan2014translating,simonyan2014,tran2014c3d}. However,
there has also been a stream of research on using recurrent neural networks
(RNN) for modeling video clips. For instance, in \cite{Srivastava2015},
Srivastava et al.  propose to use long short-term memory units to extract video
features. Ranzato et al. in \cite{Ranzato2014} also models a video clip with an
RNN, however, after vector-quantizing image patches of the video clip. In contrast to 
other approaches such as \cite{donahue2014long}, which have explored CNN-RNN coupled 
models for video description, here we use an attention mechanism, use a 3-D CNN and focus
on open-domain video description.

%\alert{TBD}

\section{Experiments}
\label{sec:exp}

We test the proposed approaches on two video-description corpora:
Youtube2Text~\cite{Chen2011} and DVS~\cite{Torabi2015}. Implementations are available at \url{https://github.com/yaoli/arctic-capgen-vid}.
%
%\vspace{-.15in}
\subsection{Datasets}
\paragraph{Youtube2Text}
The Youtube2Text video corpus~\cite{Chen2011} is well suited for training and
evaluating an automatic video description generation model. The dataset has
1,970 video clips with multiple natural language descriptions for each video
clip. In total, the dataset consists of approximately 80,000 video /
description pairs, with the vocabulary of approximately 16,000
unique words. The dataset is open-domain and covers a wide range of topics including sports,
animals and music. Following \cite{venugopalan2014translating}, we split the dataset into a training set of
1,200 video clips, a validation set of 100 clips and a test set consisting of
the remaining clips.
%
%\vspace{-.15in}
\paragraph{DVS}
The DVS dataset was recently introduced in \cite{Torabi2015} with a much larger
number of video clips and accompanying descriptions than the existing
video/description corpora such as Youtube2Text. It contains video
clips extracted from 92 DVD movies along with semi-automatically transcribed
descriptive video service (DVS) narrations. The dataset consists of 49,000 video
clips covering a wide variety of situations. We follow the standard split of the
dataset into a training set of 39,000 clips, a validation set of 5,000 clips and
a test set of 5,000 clips, as suggested by \cite{Torabi2015}.
%
%\vspace{-.15in}
\paragraph{Description Preprocessing}
We preprocess the descriptions in both the Youtube2Text and DVS datasets with
{\tt wordpunct\_tokenizer} from the NLTK toolbox.\footnote{
    \url{http:/s/www.nltk.org/index.html}}. We did not do
any other preprocessing such as lowercasing and rare word elimination. After 
preprocessing, the numbers of unique words were 15,903 for Youtube2Text
and 17,609 for DVS Dataset.
\vspace{-.15in}
\paragraph{Video Preprocessing}
To reduce the computational and memory requirement,
 we only consider the first 240 frames of each video \footnote{
When the video clip has less than 240 frames, we pad the video with all-zero
frames to make it into 240-frame long.}
For appearance features, (trained) 2-D
\textit{GoogLeNet}~\cite{szegedyL} CNN is used to extract
fixed-length representation (with the help of the popular 
implementation in Caffe~\cite{jia2014caffe}). 
Features are extracted from the {\tt pool5/7x7\_s1} layer. We select 26
equally-spaced frames out of the first 240 from each video
and feed them into the CNN to obtain a 1024 dimensional frame-wise feature vector. 
We also apply the spatio-temporal 3-D CNN (trained as described in Sec. \ref{sec:training}) in order to extract local motion information\footnote{
    We perturb each video along three axes to form random crops by taking multiple $15\times 15
    \times 120$ cuboids out of the original $20\times 20 \times 120$ cuboids,
    and the final representation is the average of the representations from
    these perturbed video clips.}.
When using 3-D CNN without temporal attention, we simply use the
2500-dimensional activation of the last fully-connection layer.
When we combine the 3-D CNN with the temporal attention mechanism, we leverage
the last convolutional layer representation leading to 26 feature vectors of
size 352. Those vector are contatenated with the 2D CNN
features resulting in 26 feature vectors with 1376 elements.

\begin{table*}[ht]
\centering
\caption{Performance of different variants of the model on the Youtube2Text and DVS datasets. 
% Note that \cite{venugopalan2014translating} used the convolutional neural network 
% from \cite{krizhevsky2012} which has lower object recognition performance
 % than the one we used from \cite{szegedyL}.
% CPAL removed the above as it is better to state such things in the body text 
 }
\label{tab:result}
\begin{tabular}{l||cccc||cccc}
 & \multicolumn{4}{c||}{Youtube2Text} & \multicolumn{4}{c}{DVS} \\
  Model                        & {\small BLEU} & {\small METEOR} & {\small CIDEr}  & {\small Perplexity}   & 
  {\small BLEU} & {\small METEOR} & {\small CIDEr}  & {\small Perplexity}   
  \\ 
\hline
\hline
Enc-Dec (Basic) & 0.3869     & 0.2868 & 0.4478 & 33.09       
 & 0.003      & 0.044  & 0.044  & 88.28       \\
+ Local (3-D CNN) & 0.3875     & 0.2832 & 0.5087 & 33.42       
 & 0.004      & 0.051  & 0.050  & 84.41       \\
+ Global (Temporal Attention) & 0.4028     & 0.2900 & 0.4801 & 27.89       
 & 0.003     & 0.040  & 0.047  & 66.63       \\
+ Local + Global & \bf 0.4192     & \bf 0.2960 & \bf 0.5167 & \bf 27.55       
 & \bf 0.007     & \bf 0.057  & \bf 0.061  & \bf 65.44       \\
\hline
Venugopalan~\textit{et al.}~\cite{venugopalan2014translating} & 0.3119     & 0.2687 & -      & -           
& -     & - & -      & -           \\
+ Extra Data (Flickr30k, COCO)                                 & 0.3329     & 0.2907 & -      & -           
& -     & - & -      & -           \\
\hline
Thomason ~\textit{et al.}~\cite{thomason2014} & 
0.1368     & 0.2390 & -      & -           
& -     & - & -      & -           \\
\end{tabular}
\end{table*}

\subsection{Experimental Setup} %Was Settings

\paragraph{Models}

% CPAL - I think it would help to give a quick recap with a little more detail as to what each model being used here consists of
% -- and how they relate to specific sections, some of the old text (below) could be merged back in perhaps
% I think this will help reviewers more quickly understand the table of results

We test four different model variations for video description generation based on the
underlying encoder-decoder framework, with results presented in Table~\ref{tab:result}. 
\emph{Enc-Dec (Basic)} denotes a baseline incorporating neither local nor global
temporal structure. Is it based on an
encoder using the 2-D GoogLeNet CNN \cite{szegedyL} as discussed in Section~\ref{sec:cnn} 
and the LSTM-based decoder outlined in Section
\ref{sec:decoder}. \emph{Enc-Dec + Local} incorporates local temporal structure via the integration of our
proposed 3-D CNN features (as outlined in Section \ref{sec:3dconv}) with the
2-D GoogLeNet CNN features as described above. \emph{Enc-Dec + Global} adds
the temporal attention mechanism of Section \ref{sec:attention}. Finally,
\emph{Enc-Dec + Local + Global}
incorporates both the 3-D CNN and the temporal attention mechanism into the
model. 
%%% This is what I had written, but then noticed that the previous section was giving the sufficient details
%temporal structure. 
%Is it based on an
%encoder using the 2-D GoogLeNet CNN \cite{szegedyL} as discussed in Section~\ref{sec:cnn} 
%and the LSTM-based decoder outlined in Section~\ref{sec:decoder}.
%We compare it with our model,
%\emph{Enc-Dec + Local + Global},
%which incorporates both the 3-D CNN of Section \ref{sec:3dconv} and the temporal attention mechanism
%of Section~\ref{sec:attention}.  
%We also compare with exploiting either the local or global structure only.
%\emph{Enc-Dec + Local} incorporates local temporal structure by concatenating to the
%representation of Eq.~\ref{eq:simple_avg} the fully connected layer (FC4) of the 3-D CNN (see Fig.~\ref{fig:architecture}). 
%emph{Enc-Dec + Global} on the other hand doesn't use at all the 3-D CNN features. 
All models
otherwise use the same number of temporal features $\vv_i$.
These experiments will allow us to investigate
whether the contributions from the proposed approaches are complimentary
and can be combined to further improve performance.

%We test three different configurations of the automatic video description
%generation model based on the encoder-decoder. 
%The first model does not incorporate any temporal structure.
 %CPAL dont need the citation here, as it is not exactly the same
 % ~\cite{venugopalan2014translating}. 
%We add to this model either the temporal attention mechanism or the spatio-temporal
%convolutional neural network to test each of the two proposed approaches. Last,
%we build and evaluate a model that has both the temporal attention mechanism and
%the spatio-temporal convolutional neural network to investigate whether the
%contributions from the proposed approaches are orthogonal and can be combined to
%improve the performance even further.

\paragraph{Training}\label{sec:training}
For all video description generation models, we estimated the parameters by
maximizing the log-likelihood:
\begin{align*}
    \mathcal{L}(\TT) = \frac{1}{N} \sum_{n=1}^N \sum_{i=1}^{t_n} \log p(y_i^n \mid
    y_{<i}^n, \vx^n, \TT),
\end{align*}
where there are $N$ training video-description pairs $(\vx^n, y^n)$, and each
description $y^n$ is $t_n$ words long. 

We used Adadelta~\cite{Zeiler-2012} with the gradient computed by the
backpropagation algorithm. We optimized the hyperparameters (e.g. number of LSTM units and the word
embedding dimensionality) using random search to maximize the log-probability of
the validation set.
\footnote{
    Refer to the Supplementary Material for the selected hyperparameters.
} 
Training continued until the validation log-probability stopped increasing for
5,000 updates.
As mentioned earlier in Sec.~\ref{sec:3dconv}, the 3-D CNN was trained on activity
recognition datasets.
Due to space limitation, details regarding the training and evaluation of the 3-D CNN
on activity recognition datasets are provided in the Supplementary Material.
%We evaluated the trained 3-D CNN on HMDB51 (split 1) and
%confirmed that the trained model is reasonable (3\% lower test accuracy than the best temporal convolutional neural network from \cite{simonyan2014}).
%Note that the HMDB51 dataset is temporally trimmed and \cite{simonyan2014}
%took advantage of averaging scores of 250 video samples of 15 frames, plus cropping and flipping data augmentation, out of each video clip when computing the class scores. We only used 1 sample of 240 frames for each video clip.

\paragraph{Evaluation}
We report the performance of our proposed method using test set perplexity
and three model-free automatic evaluation metrics. These are
BLEU~\cite{papineni2002bleu}, METEOR~\cite{meteor:2014}
and CIDEr~\cite{vedantam2014cider}. We use the
evaluation script prepared and introduced in \cite{chen2015microsoft}. 

\subsection{Quantitative Analysis}
\label{sec:quantitative}

In the first block of Table~\ref{tab:result}, we present the performance of
the four different variants of the model
using all four metrics: BLEU, METEOR, CIDEr and perplexity. Subsequent lines in the table give comparisons with prior work.
The first three rows (Enc-Dec (Basic), +Local and +Global), show that it is
generally beneficial to exploit some type of temporal structure
underlying the video. Although this benefit is most evident with perplexity
(especially with the temporal attention mechanism exploiting global temporal
structure), we observe a similar trend with the other
model-free metrics and across both Youtube2Text and DVS
datasets.

We observe, however, that the biggest gain can be achieved by letting the model
exploit {\it both} local and global temporal structure (the fourth row in
Table~\ref{tab:result}). We observed this gain consistently across both datasets
as well as using all four automatic evaluation metrics.

\begin{figure*}[ht]
\centering
\includegraphics[width=\textwidth]{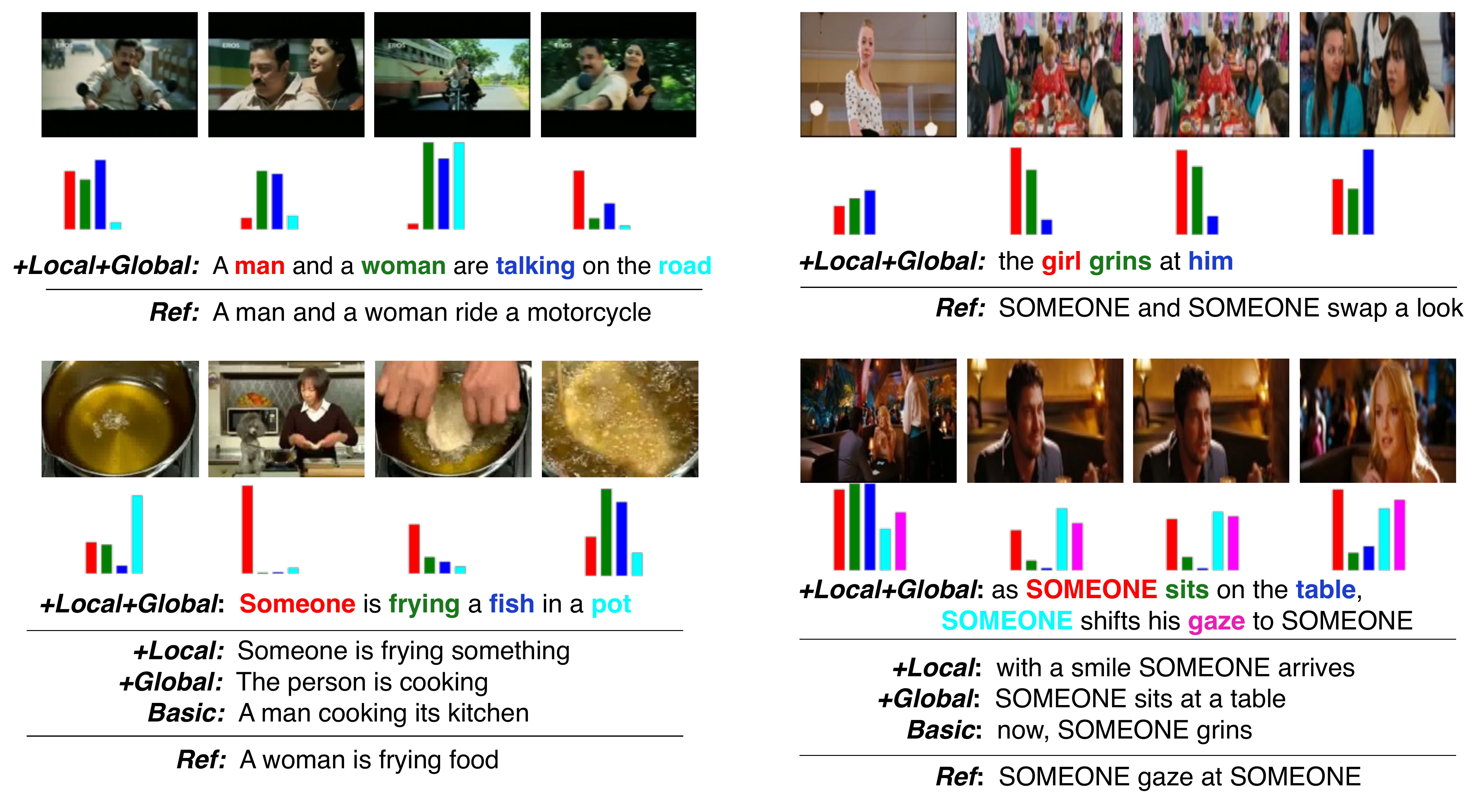}
\caption{Four sample videos and their corresponding generated and ground-truth
descriptions from Youtube2Text (Left Column) and DVS (Right Column). The bar plot under each frame corresponds
to the attention weight $\alpha_i^t$ for the frame when the corresponding word
(color-coded) was generated. From the top left panel, we can
see that when the word ``road'' is about to be generated, the model focuses
highly on the third frame where the road is clearly visible. Similarly, on the
bottom left panel, we can see that the model attends to the second frame
when it was about to generate the word ``Someone''. The bottom row includes
alternate descriptions generated by the other model variations.}
\label{fig:vis_samples}
\end{figure*}

\subsection{Qualitative Analysis}

Although the model-free evaluation metrics such as the ones we used in this
paper (BLEU, METEOR, CIDEr) were designed to reflect the agreement level
between reference and generated descriptions, it is not intuitively
clear how well those numbers (see Table~\ref{tab:result}) reflect the
quality of the actual
generated descriptions. Therefore, we present some of the video clips and
their corresponding descriptions, both generated and reference, from the
test set of each dataset. Unless otherwise labeled, the visualizations in this section are from the best model
which exploits both global and local temporal structure (the fourth row of
Table~\ref{tab:result}). 

In Fig.~\ref{fig:vis_samples}, two video clips from the test set of
Youtube2Text are shown. We can clearly see that the generated descriptions
correspond well with the video clips. 
In Fig.~\ref{fig:vis_samples}, we show also two sample video clips from the
DVS dataset. Clearly, the model does not perform as well on the DVS dataset as
it did on Youtube2Text, which was already evident from the quantitative analysis
in Sec.~\ref{sec:quantitative}. However, we still observe that the model often
focuses correctly on a subset of frames according to the word to be generated.
For instance, in the left pane, when the model is about to generate the second
``SOMEONE'', it focuses mostly on the first frame. Also, on the right panel, the
model correctly attends to the second frame when the word ``types'' is about to
be generated. As for the 3-D CNN local temporal features, we see that they allowed
to correctly identify the action as ``frying'', as opposed to simply ``cooking''.

More samples of the video clips and the generated/reference descriptions can be
found in the Supplementary Material, including visualizations from the global temporal 
attention model alone (see the third row in Table~\ref{tab:result}).

\section{Conclusion}
In this work, we address the challenging problem of producing natural
language descriptions of 
videos. %In addition to the frame-wise appearance information, 
%we identify the importance of capturing both local and global temporal structure.
We identify and underscore the importance of capturing both local and global temporal structure 
in addition to frame-wise appearance information.
To this end, we propose a novel 3-D convolutional neural network that is
designed to capture local fine-grained motion information from consecutive frames. 
In order to capture global temporal structure, we propose the use of
a temporal attentional mechanism that learns the ability to focus on
subsets of frames. 
%learn a distribution of importance weights over frames, one per word,  with the capability to take into account global temporal information from non-consecutive frames. On the other hand, 
Finally, the two proposed approaches fit naturally together into an encoder-decoder neural 
video caption generator. 

We have empirically validated each approach on both Youtube2Text and DVS datasets on 
four standard evaluation metrics. 
Experiments indicate that models using either approach improve over the
baseline model. Furthermore, 
combining the two approaches gives the best performance. 
In fact, we achieved the state-of-the-art results on Youtube2Text with the combination.

Given the challenging nature of the task, we hypothesize that the performance
on the DVS dataset could be significantly improved by incoporating another
recently proposed dataset  \cite{Rohrbach2015} similar to the DVS data used here. In addition, we
have some preliminary experimental results that indicate that further
performance gains are possible by leveraging image caption  generation datasets
such as MS COCO~\cite{chen2015microsoft} and Flickr~\cite{hodosh2013framing}.
We intend to more fully explore this direction in future work.

 \section*{Acknowledgments} 
 The authors would like to thank the developers of Theano ~\cite{bergstra+al:2010-scipy,Bastien-Theano-2012}. We acknowledge the support of the following organizations for research funding and computing support: NSERC, FQRNT, Samsung, Calcul Quebec, Compute Canada, the Canada Research Chairs and CIFAR.

{\small
\bibliographystyle{ieee}
\bibliography{camera_ready}
}

\onecolumn
\section{Details of experiments}

\subsection{3-D CNN}

\begin{figure*}[!ht]
\center
\begin{minipage}{0.65\textwidth}
\includegraphics[width=\columnwidth]{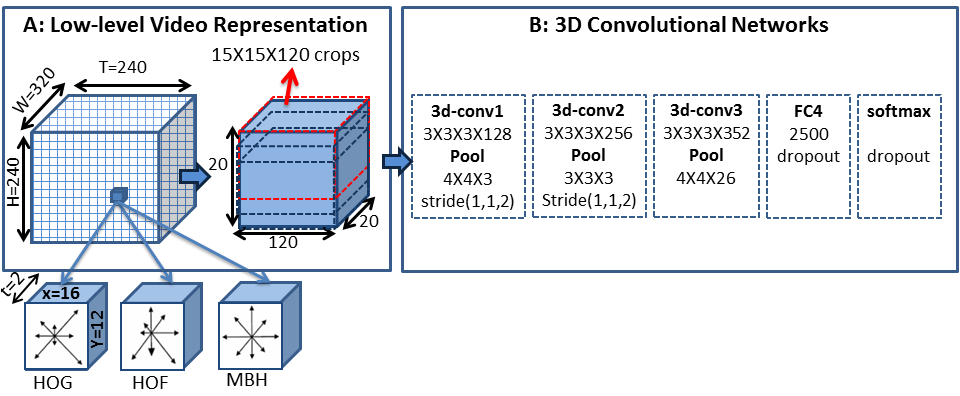}
\end{minipage}
\hspace{1mm}
\begin{minipage}{0.33\textwidth}
\caption{\small Illustration of the spatio-temporal convolutional neural
    network (3-D CNN).
    %In a 3DConv layer,
%convolution dimensions are $width \times height \times time \times channels$.
%For 3-D pooling dimensions are $width \times height \times time$. The stride
%dimensions for pooling of the first and second convolution layers are $width
%\times height \times time$.
This network is trained for activity recognition. Then, only the convolutional
layers are involved when generating video descriptions.
%In training, random crops (in red) are extracted and
%input to the first convolutional layer.
}
\label{fig:architecture}
\end{minipage}
\end{figure*}

The 3-D CNN architecture is specified in Figure~\ref{fig:architecture}.
Our model is composed of three 3-D convolutional layers, using 3$\times$3$\times$3 kernels. The number of output features after the different convolutions is given in Figure~\ref{fig:architecture}.  Each convolutional layer is followed by a rectified linear
activations (ReLU) and local max-pooling.
After the convolutions, a fully-connected layer (dimension 2500, with ReLU activation) is applied, followed by a Softmax layer.
A dropout of 0.5 is applied on those last two layers.\\

Multitask learning is used to train the model on three human
activity recognition datasets: UCF101~\cite{Khurram2012} with 13320 Youtube videos and 101 various human activity classes, HMDB51~\cite{Kuehne2011} with 3700 videos and 51 various human activity classes, and a random subset
of Sports-1M dataset~\cite{KarpathyCVPR14} using 50,000 videos that have 487 sports labels.

We trained the 3-D CNN using stochastic gradient descent with a  momentum of 0.7.
Learning rate is initially set to 0.1 and then is decreased, using the
following scheme 0.05, 0.02, 0.01, each time the validation cost stagnate.
At each iteration a minibatch of size 48 is constructed by sampling uniformly 
all 3 human activity datasets.
We perturb each video along three axes to encourage the model to learn invariant feature representation, we take random crops of size 15$\times$15$\times$120 cuboids out of the original 20$\times$20$\times$120 cuboids. Video are also randomly flipped.\\

Despite our interest in video-description, we validate
that our model obtains reasonable performances on the activity recognition task
 using HMDB51 (split 1)  and UCF101.
On HMDB (split 1) our model achieves an accuracy of $52.3\%$, our result is $3\%$ lower than the best motion-based single model (temporal-based CNN of~\cite{simonyan2014}).
On UCF-101, our 3-D CNN obtain an accuracy of $76.49\%$. While the temporal-based
CNN~\cite{simonyan2014} achieves $83.7\%$, our model outperforms other single 3-D convolution based approaches such as C3D ($72.29\%$)~\cite{tran2014c3d}
and slow-fusion convnet ($65.4\%$)~\cite{karpathy2014large}.

\subsection{Encoder-Decoder Model Training}

Hyperparameters reflects the learning capacity of the models. We have made sure each type of models have been sufficient explored in their hyperparameters.
The model selections on both Youtube2Text and DVS are performed by random search~\cite{bergstra2012random}.
There are four types of models being trained:
\begin{itemize}
\item Basic Enc-Dec
\item Basic Enc-Dec + Local (3-D ConvNet)
\item Basic + Global (Temporal Attention)
\item Basic + Local + Global
\end{itemize}

For each of four types of models, we performed 50 experiments with random search on the
critical hyperparameters. Each of the 50 experiments is associated with a specific
hyperparameter setup. And the 50 setups are shared cross all four types of models.
The critial hyperparameters experimented are:
\begin{itemize}
\item the dimensionality of word embedding, in the range of [100, 1000]
\item the dimensionality of LSTM hidden/memory states, in the range of [100,3000]
\item dropout, either use or not used, decided at random.
\end{itemize}

The same procedure is used on both Youtube2Text and DVS datasets.
\begin{table}[h]
\caption{Hyperparameters of best models on Youtube2Text.}
\small
\centering
\begin{tabular}{l l l l}
\hline
model                                            &emb   &lstm    & dropout \\
\hline
Basic Enc-Dec                                   & 211  & 1096   & True    \\
Basic Enc-Dec + Local (3-D ConvNet)             & 161  & 1292   & True    \\
Basic + Global (Temporal Attention)             & 476  & 2231   & True    \\
Basic + Local + Global                          & 454  & 1714   & True    \\
\hline
\end{tabular}
\end{table}

\begin{table}[h]
\caption{Hyperparameters of best models on DVS.}
\small
\centering
\begin{tabular}{l l l l}
\hline
model                                    &emb   &lstm    & dropout \\
\hline
Basic Enc-Dec                               & 345  & 1014   & True    \\
Basic Enc-Dec + Local (3-D ConvNet)         & 512  & 2560   & True    \\
Basic + Global (Temporal Attention)                      & 656  & 1635   & True    \\
Basic + Local + Global       & 454  & 1714   & True    \\
\hline
\end{tabular}
\end{table}

\section{Inspecting the learned soft-attention coefficients $\alpha$}
We illustrate the caption generation process of 
the proposed soft-attentional models trained with Basic+Global v.s. 
trained with Basic+Local+Global, with a dynamic $\alpha$ on frames for each 
 word in the generated caption. 

The bar chart shows the magnitude of $\alpha$. The generated caption is shown on the left.
Each generated word corresponds to an $\alpha$ vector, show in the same row.
Each bar corresponds to a particular frame on the very top of the figure, organized
sequentially. Within
the same row, the height of the bar shows the importance of its corresponding frame
in generating that word.
20 frames are shown for better visibility.
\subsection{Caption generation and $\alpha$ visualization on Youtube2Text testset}
\begin{figure*}[ht]
\centering
\includegraphics[scale=0.37]{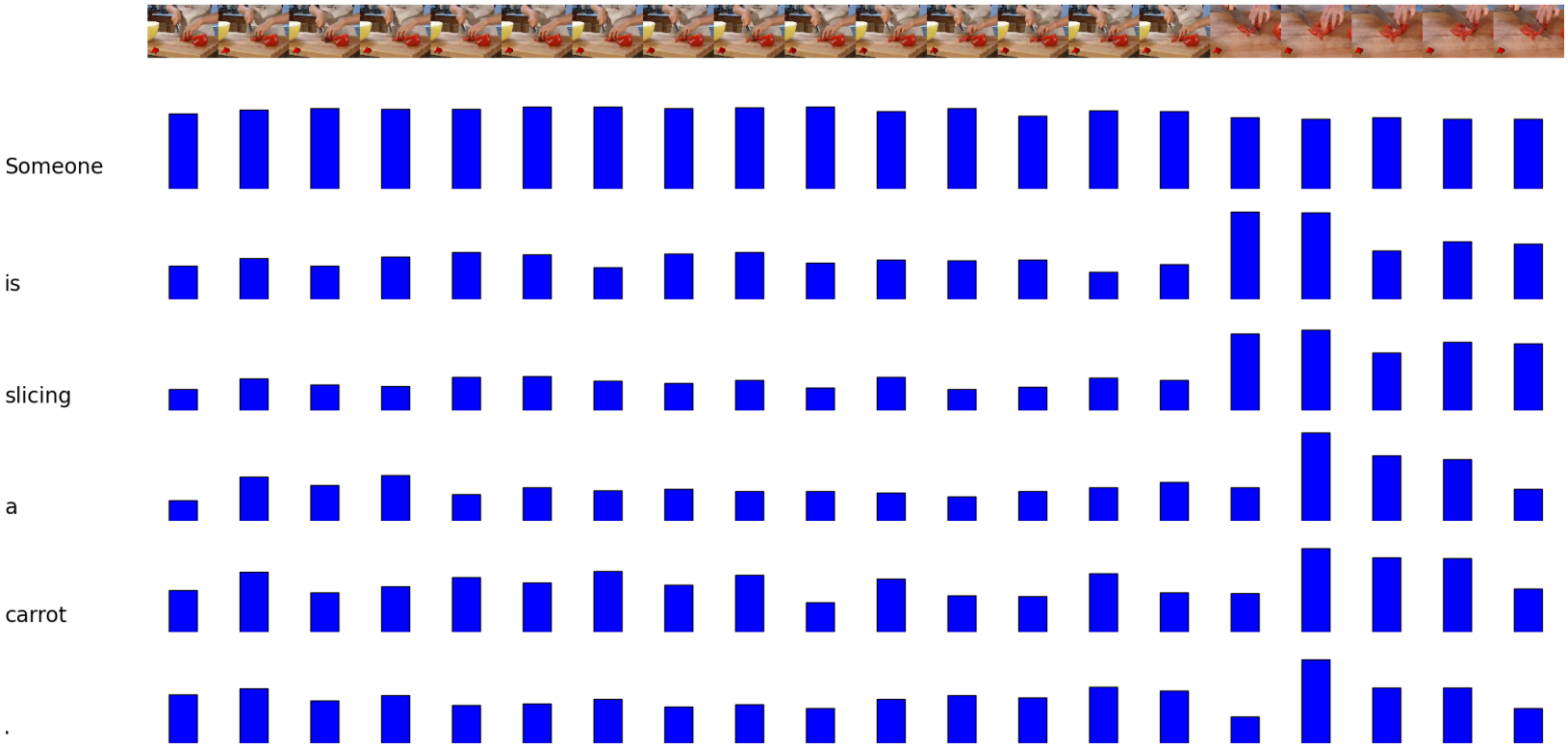}
\caption{
Model type: Basic + Global. 
Model shifts its attention across frames to generate a caption. 
The bar char shows the magitude of $\alpha$, sum to 1 row-wise, the higher the bar, 
the bigger the magnitude.}\label{y2t:1}
\end{figure*}
\begin{figure*}[ht]
\centering
\includegraphics[scale=0.37]{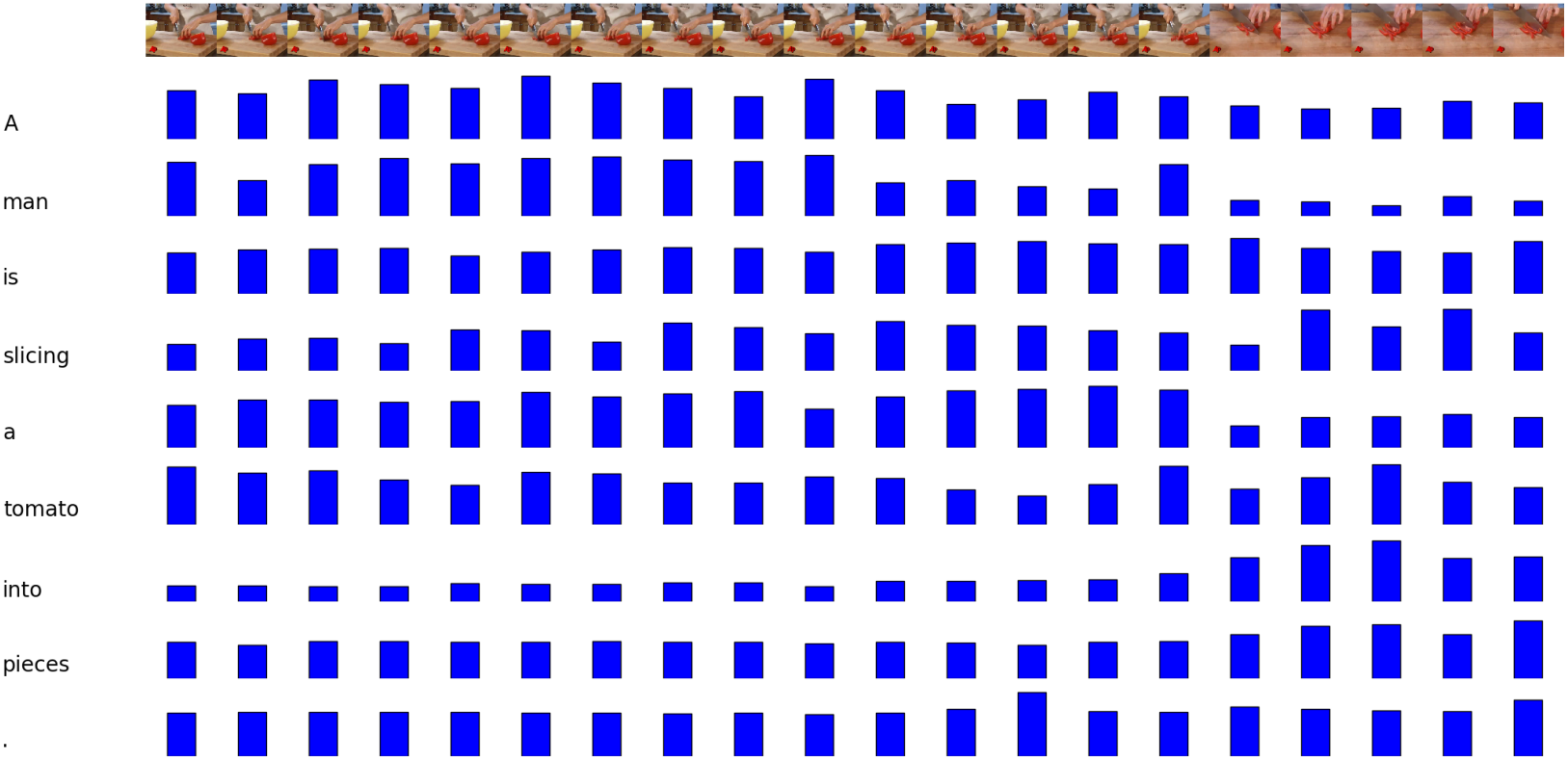}
\vspace{2cm}
\caption{
Model type: Basic + Local + Global.
Model shifts its attention across frames when generating the caption.
The bar char shows the magitude of $\alpha$, sum to 1 row-wise, the higher the bar,
the bigger the magnitude. It is doing a better job at guessing the object being chopped
, compared with Figure \ref{y2t:1}.}
\end{figure*}

\begin{figure*}[ht]
\centering
\includegraphics[scale=0.37]{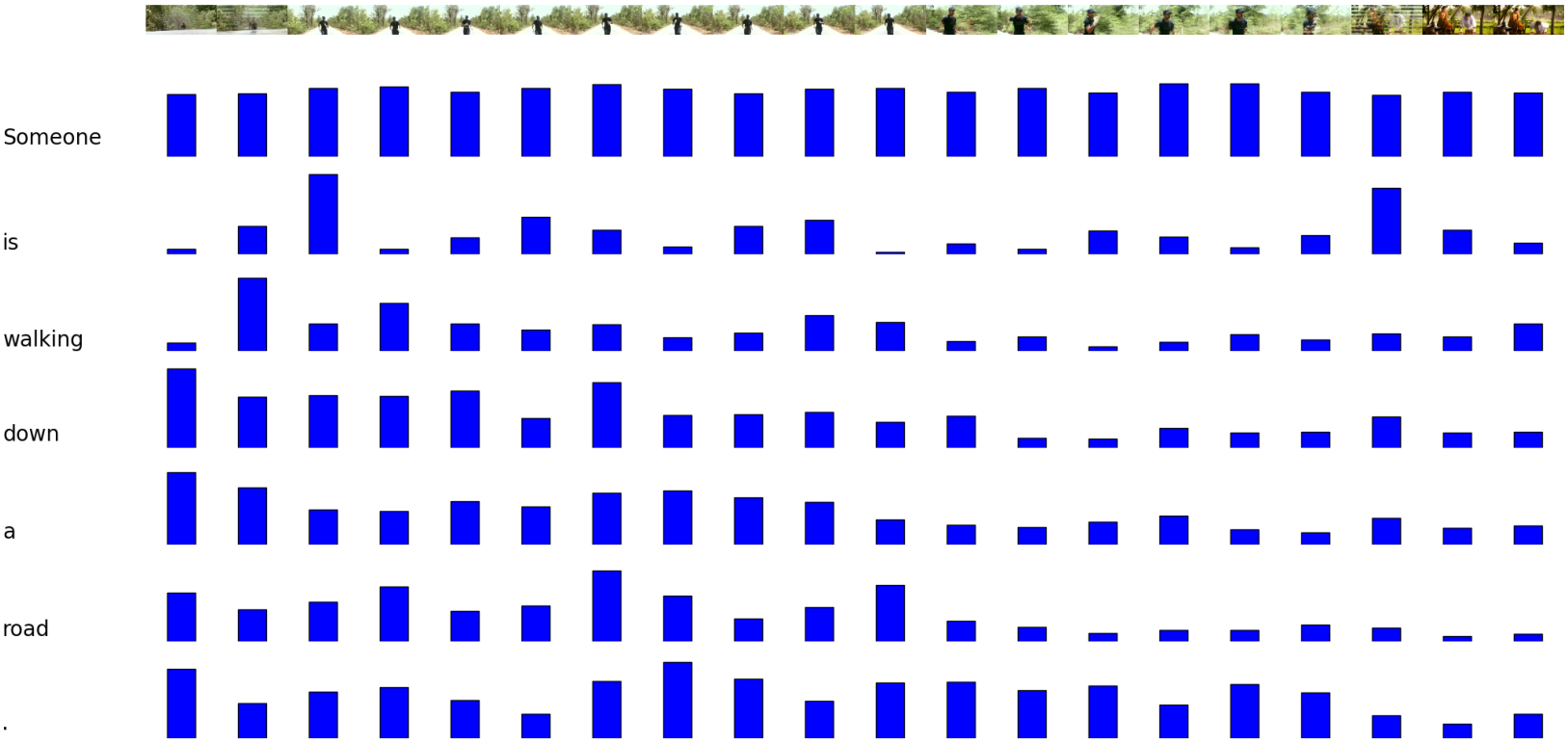}
\caption{
Model type: Basic + Global. 
Model shifts its attention across frames to generate a caption. 
The bar char shows the magitude of $\alpha$, sum to 1 row-wise, the higher the bar, 
the bigger the magnitude.} \label{y2t:2}
\end{figure*}
\begin{figure*}[ht]
\centering
\includegraphics[scale=0.37]{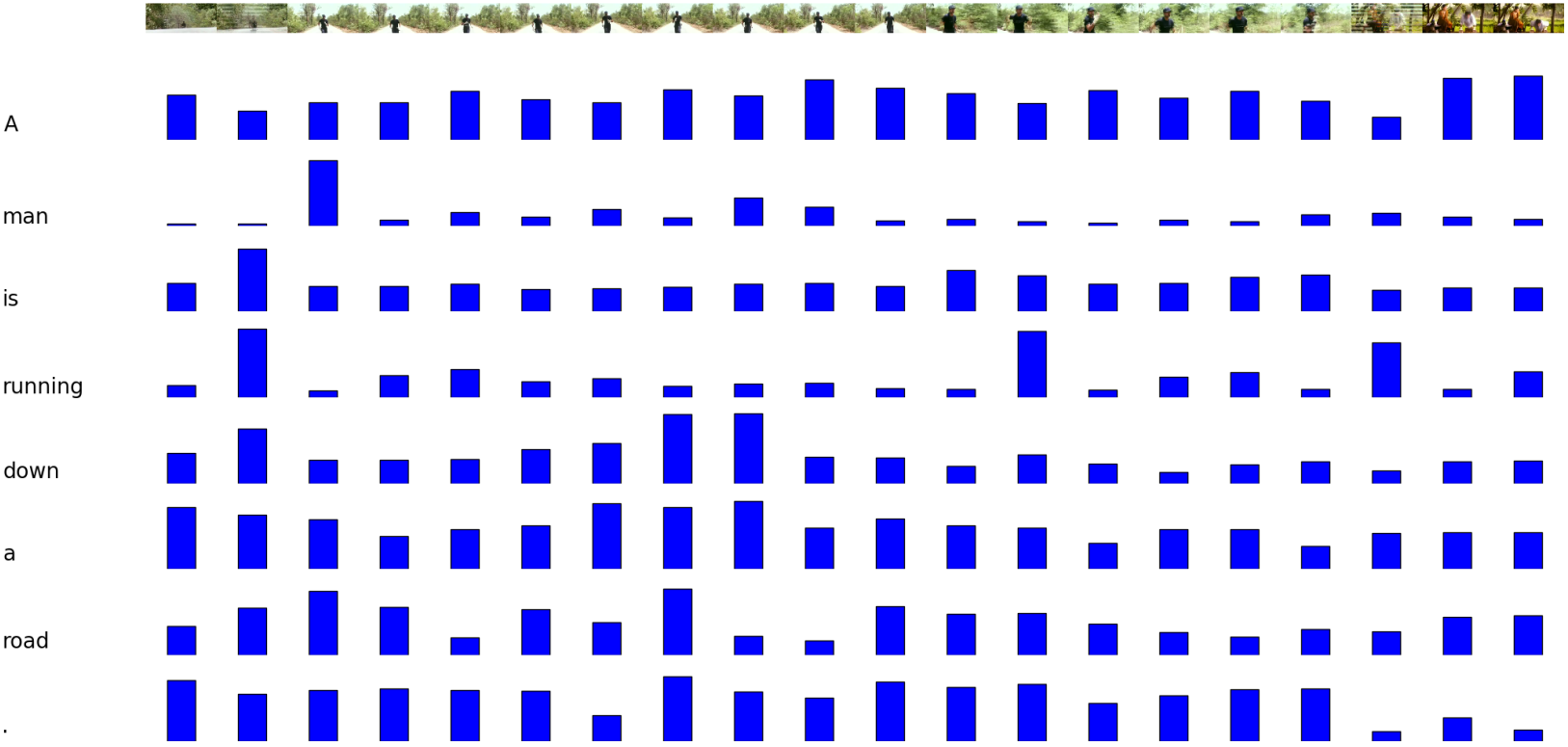}
\caption{
Model type: Basic + Local + Global.
Model shifts its attention across frames when generating the caption.
The bar char shows the magitude of $\alpha$, sum to 1 row-wise, the higher the bar,
the bigger the magnitude. The use of additional motion features offers more faithful
description of the action than the one without (``running'' v.s. ``walking'' in Figure \ref{y2t:2}).}
\end{figure*}

\begin{figure*}[ht]
\centering
\includegraphics[scale=0.37]{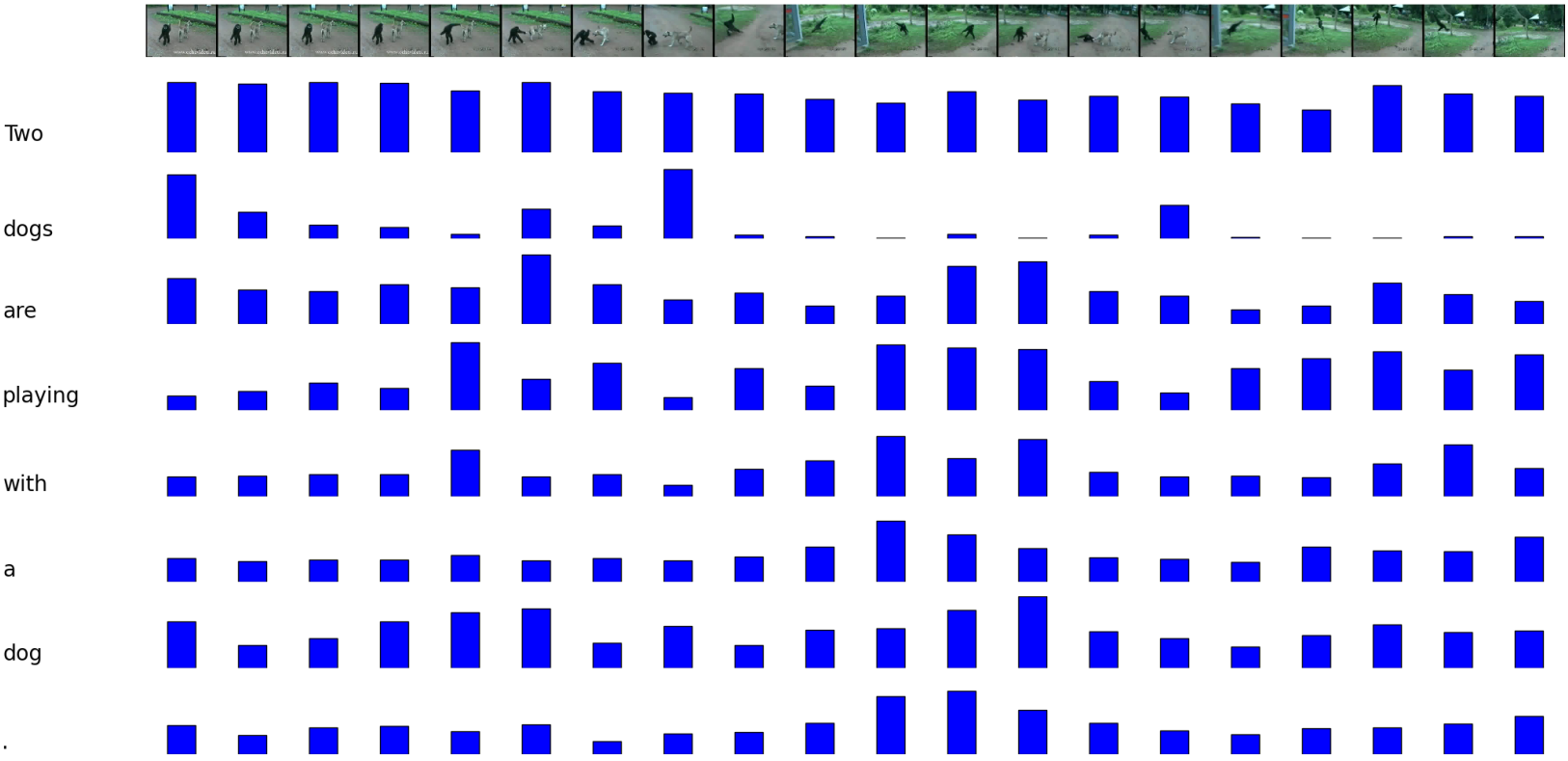}
\caption{
Model type: Basic + Global. 
Model shifts its attention across frames to generate a caption. 
The bar char shows the magitude of $\alpha$, sum to 1 row-wise, the higher the bar, 
the bigger the magnitude.} \label{y2t:3}
\end{figure*}
\begin{figure*}[ht]
\centering
\includegraphics[scale=0.37]{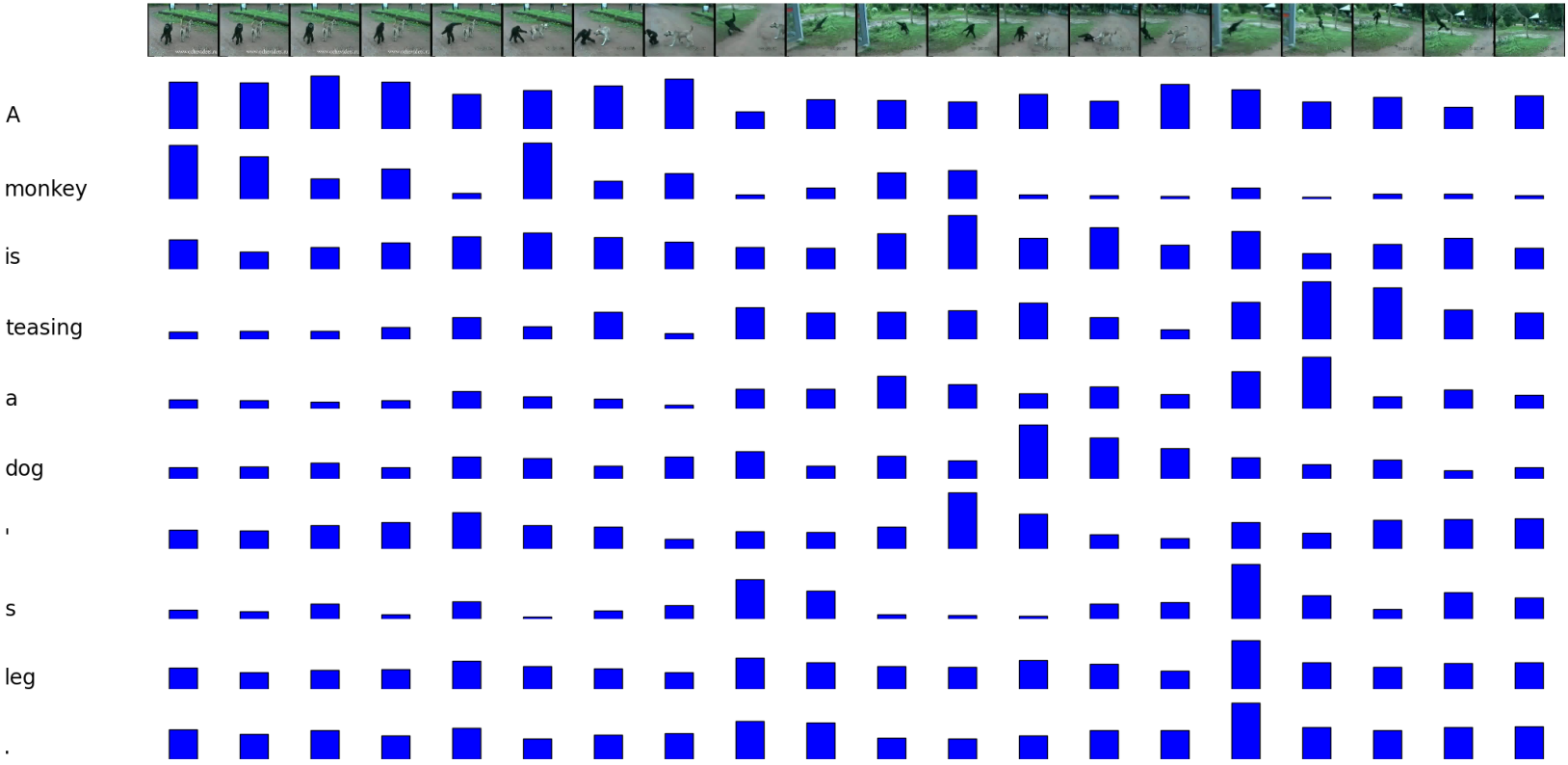}
\caption{
Model type: Basic + Local + Global.
Model shifts its attention across frames to generate a caption.
The bar char shows the magitude of $\alpha$, sum to 1 row-wise, the higher the bar,
the bigger the magnitude. $\mbox{3DConv}_{\mbox{att}}$ generates a more faithful description with a much richer content than Figure
\ref{y2t:3}. It even learns to generate a rare work ``teasing''.}
\end{figure*}

\clearpage
\subsection{Caption generation and $\alpha$ visualization on DVS testset}
This section illustrates on DVS, a much more challenging dataset. See the following figures for 
detailed explaination of soft-attention applied on videos with different properties.  

\begin{figure*}[ht]
\centering
\includegraphics[scale=0.75]{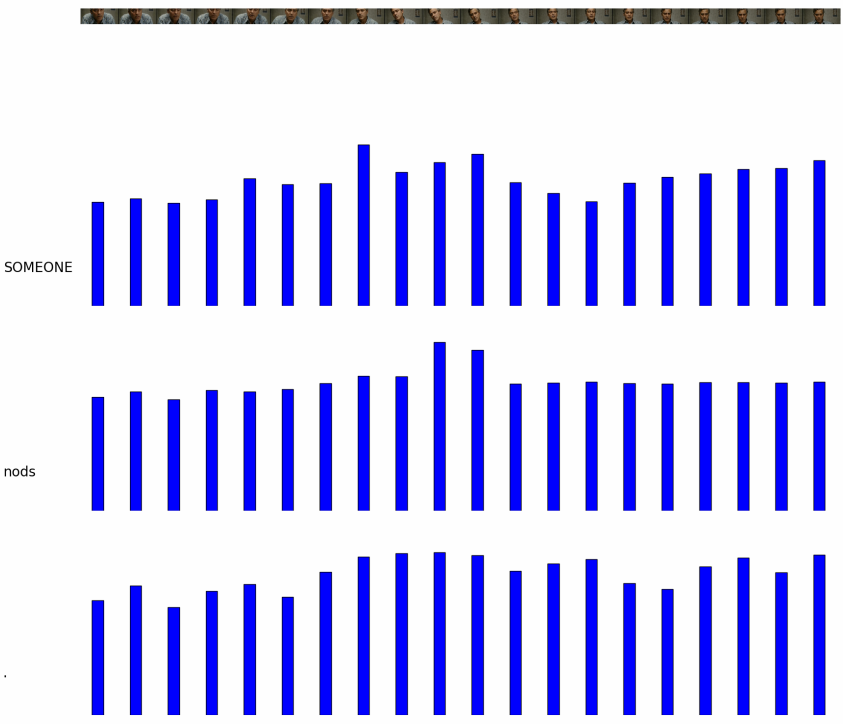}
\caption{
Model type: Basic + Global. The model tends to produce a smooth 
distribution in $\alpha$ row-wise, due to the uniformity of the scene with a slowly changing continuous shot.} \label{dvs:1}
\end{figure*}
\begin{figure*}[ht]
\centering
\includegraphics[scale=0.7]{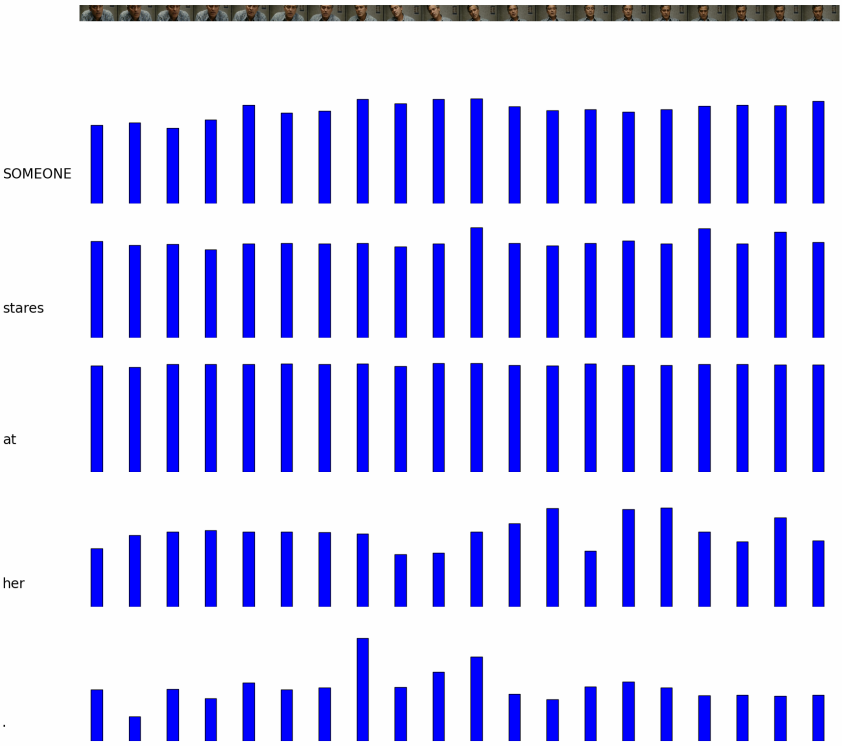}
\vspace{0cm}
\caption{
Model type: Basic + Local + Global.
The model learns a smooth $\alpha$ on the slowly changing scene.
It captures a different action from Basic + Global in Figure \ref{dvs:1}.
}
\end{figure*}

\begin{figure*}[ht]
\centering
\includegraphics[scale=0.7]{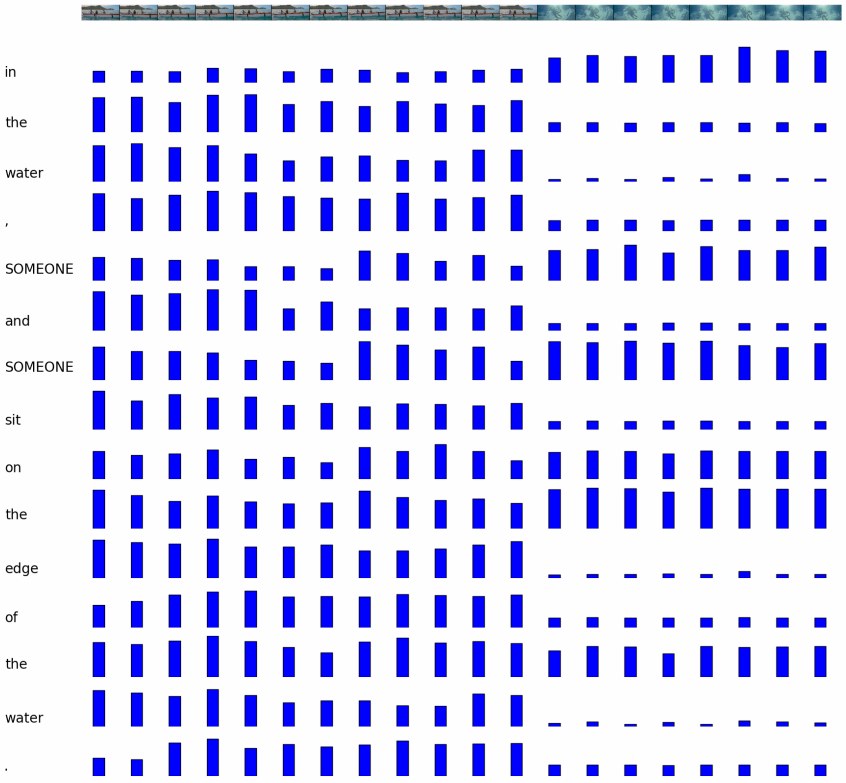}
\caption{
Model type: Basic + Global. 
$\alpha$ also reflects the sudden 
transition between two shots.} \label{dvs:2}

\end{figure*}

\begin{figure*}[ht]
\centering
\includegraphics[scale=0.7]{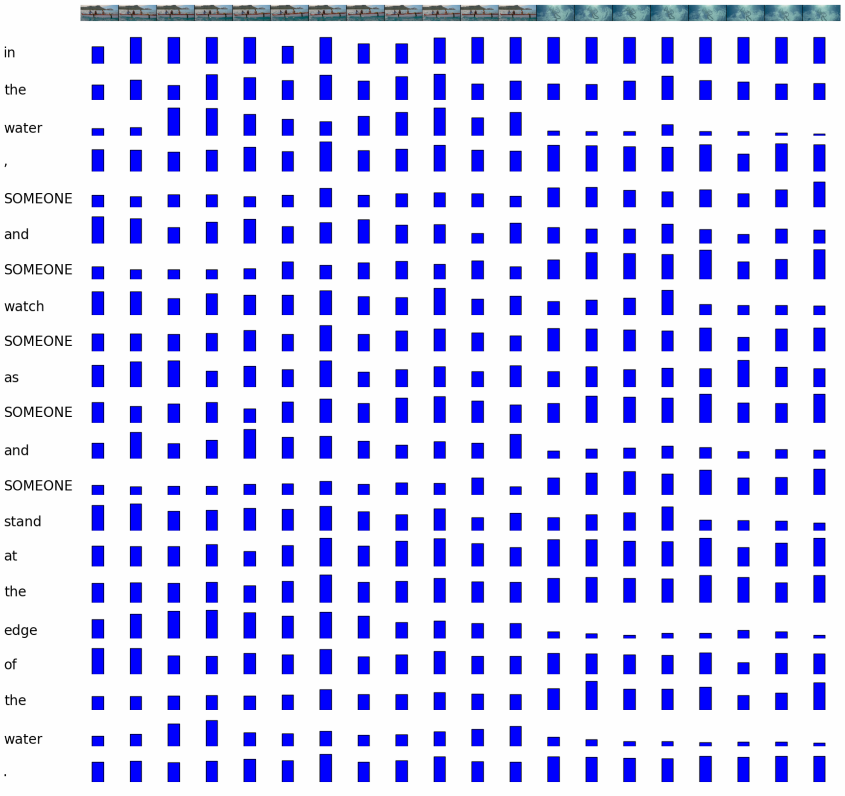}
\caption{
Model type: Basic + Local + Global. The learned model
generates a more sophiscated description than Figure \ref{dvs:2}, attempting to 
incoporate character-level interaction inside the first part of the scene.}
\end{figure*}

\begin{figure*}[ht]
\centering
\includegraphics[scale=0.75]{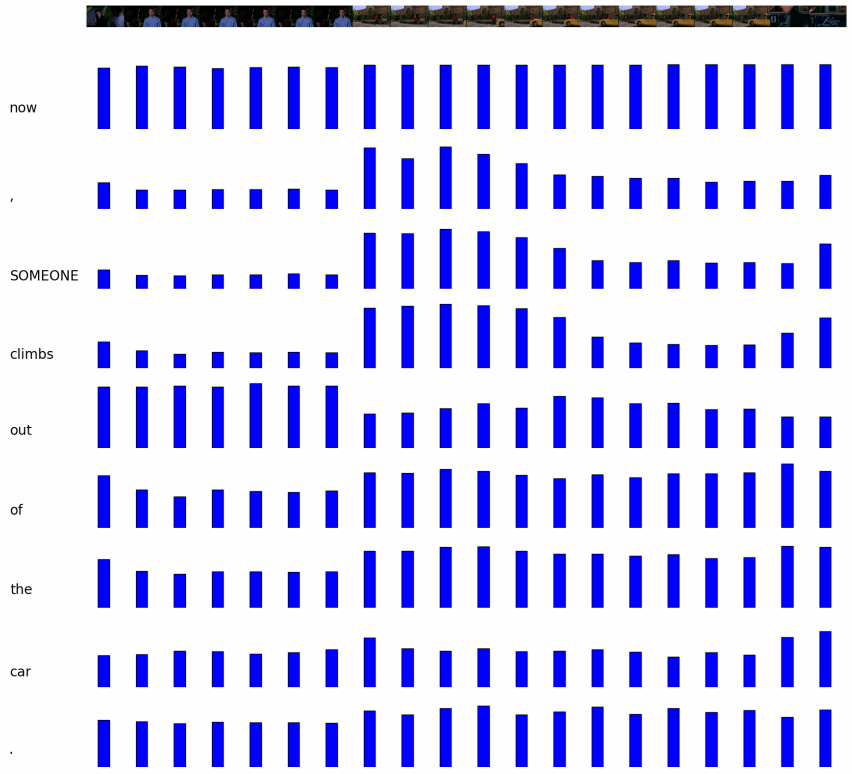}
\caption{Model type: Basic + Global. The model seems to focus on the second shot of the 
scene at the beginning, yet the part of the generated caption ``out of the car'' distributes a decent amount of its attention 
on the first scene as well. This may due to the fact that the memory of decoding LSTM already 
contains the information of almost the entire scene (two shots).} \label{dvs:3}
\end{figure*}
\begin{figure*}[ht]
\centering
\includegraphics[scale=0.75]{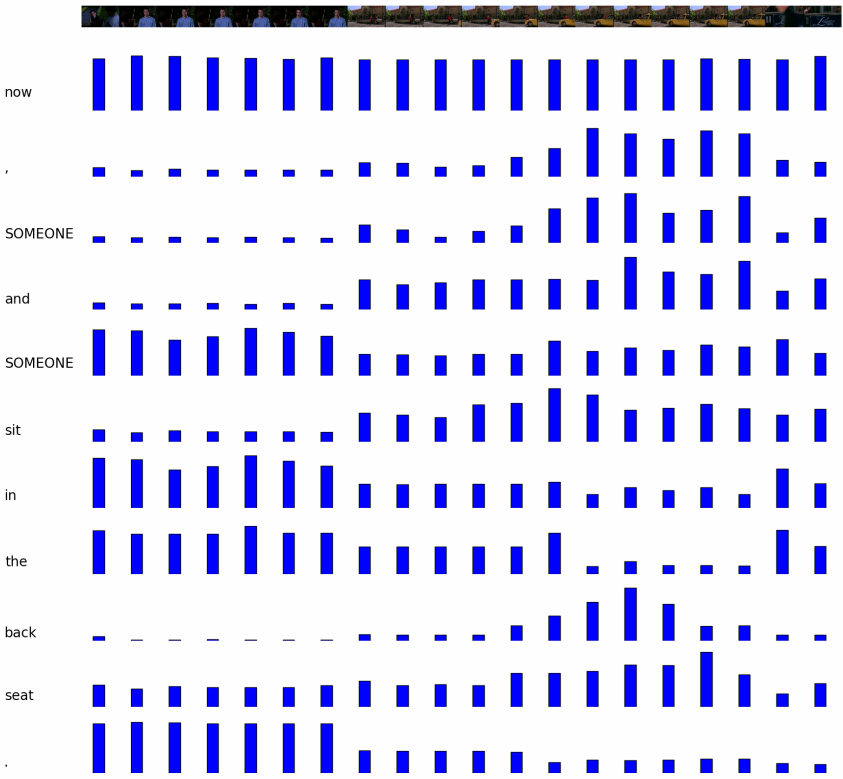}
\caption{
Model type: Basic + Local + Global. The learned model 
generates a more sophiscated description than Figure \ref{dvs:3}. The model focuses on 
the car in the second shot when generating ``sit'', ``back seat''. When generating two 
``SOMEONE'', it divides its attentio among two shots.}
\end{figure*}

\begin{figure*}[ht]
\centering
\includegraphics[scale=0.75]{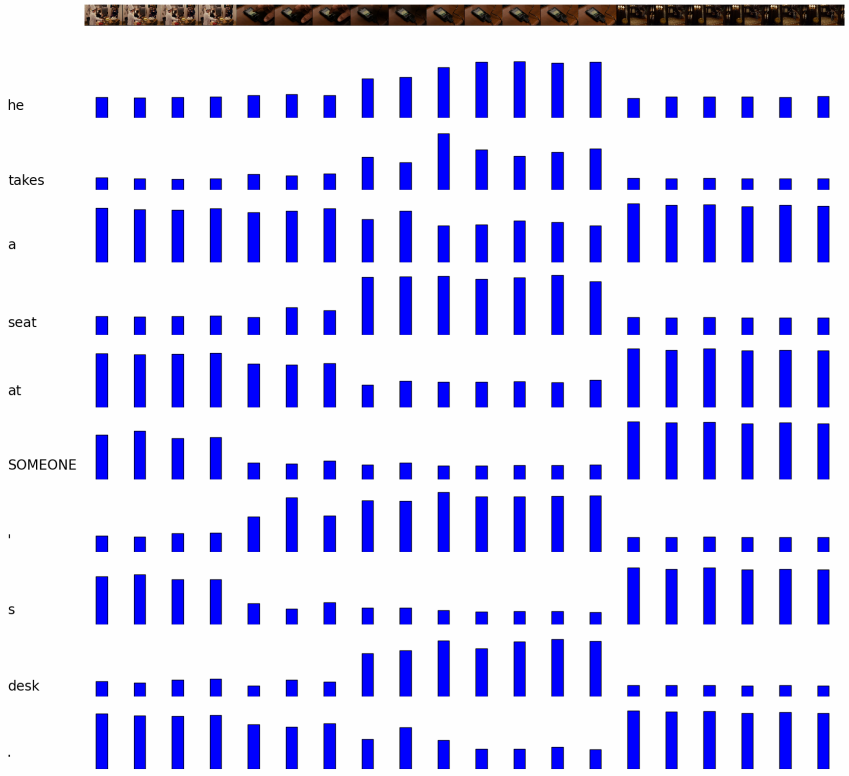}
\caption{Model type: Basic + Global. The description is 
argubly not very accurate} \label{dvs:4}
\end{figure*}
\begin{figure*}[ht]
\centering
\includegraphics[scale=0.75]{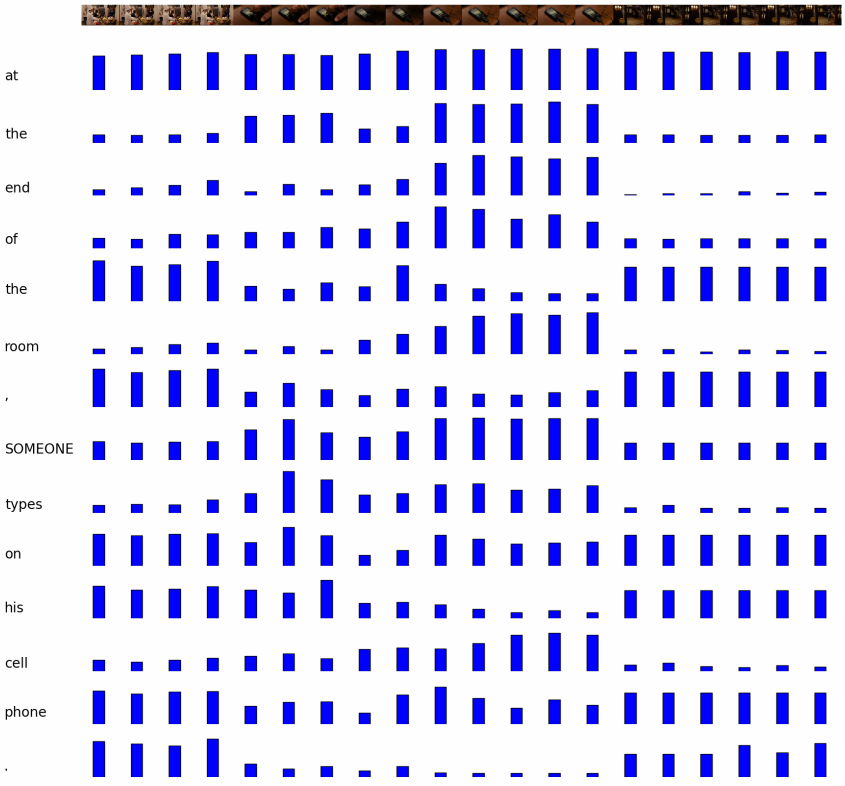}
\caption{
Model type: Basic + Local + Global. With the help of
additional features, the model successfully describes the cell phone and the room,
a much faithful description than Figure \ref{dvs:4}}
\end{figure*}

\end{document}